\documentclass[abstract=on,12pt,paper=a4]{scrartcl}

\usepackage[paper=a4paper,left=25mm,right=25mm,top=25mm,bottom=30mm]{geometry}
\usepackage[utf8]{inputenc}
\usepackage[T1]{fontenc}
\usepackage[english]{babel}

\usepackage{natbib}

\usepackage{lmodern} % Modern font
\usepackage{setspace} % Line spacing
\usepackage{tabularx}
\usepackage{arydshln}
\usepackage{microtype} % Better text spacing

\usepackage[utf8]{inputenc}
\usepackage[T1]{fontenc}
\usepackage[english]{babel}

\usepackage[font=footnotesize, labelfont=bf, format=hang, justification=raggedright, parindent=0pt]{caption}

\usepackage{xurl} % For breaking long URLs
\usepackage{amsmath, amssymb, amsthm} % Math packages
\usepackage{booktabs, tabularx, longtable} % Table formatting

\usepackage{hyperref}
\hypersetup{
    colorlinks=true,
    linkcolor=black,
    filecolor=blue,
    urlcolor=blue,
    citecolor=blue
}

\usepackage{csquotes} % Quotation marks
\usepackage{epigraph} % Fancy quotes
\usepackage{graphicx} % Image handling
\usepackage{float} % Float management
\usepackage{rotating} % Rotating elements
\usepackage{etoolbox} % For conditional logic

\usepackage{threeparttable} % Include this in your preamble
\usepackage{footmisc} % Include this for footnote sizing control
\usepackage{url} % Include this for URL formatting in footnotes % Include preamble.tex file

\begin{document}
	% Title Page
	\begin{center}
			\LARGE \doublespacing
		Comparative Insights from 12 Machine Learning Models
		in Extracting Economic Ideology from Political Text\\
		\onehalfspacing
		\large
		\vspace{40pt}
		Jihed Ncib\\
		University College Dublin\\
		\normalsize
		\vspace{20pt}
		\begin{abstract}
			\noindent
			This study conducts a systematic assessment of the capabilities of 12 machine learning models and model variations in detecting economic ideology. As an evaluation benchmark, I use manifesto data spanning six elections in the United Kingdom and pre-annotated by expert and crowd coders. The analysis assesses the performance of several generative, fine-tuned, and zero-shot models at the granular and aggregate levels. The results show that generative models such as GPT-4o and Gemini 1.5 Flash consistently outperform other models against all benchmarks. However, they pose issues of accessibility and resource availability. Fine-tuning yielded competitive performance and offers a reliable alternative through domain-specific optimization. But its dependency on training data severely limits scalability. Zero-shot models consistently face difficulties with identifying signals of economic ideology, often resulting in negative associations with human coding. Using general knowledge for the domain-specific task of ideology scaling proved to be unreliable. Other key findings include considerable within-party variation, fine-tuning benefiting from larger training data, and zero-shot's sensitivity to prompt content. The assessments include the strengths and limitations of each model and derive best-practices for automated analyses of political content.
		\end{abstract}
	\end{center}
	\newpage
	\pagenumbering{arabic}
	\doublespacing
	\hypersetup{linkcolor=blue}
	\section{Introduction}
Economic ideology is a central concept in studying political institutions. Not only does it impact political behavior and rhetoric but also policy preferences and institutional development. It drives decision-making at the individual and collective levels and encompasses a range of ideas from free-market liberalism to state interventionism \citep{downs_economic_1957, lipset_political_1960}. In addition, one cannot examine political competition and voter alignment without considering ideology, including and especially on economic policy \citep{budge_mapping_2001,huber_church-state_2011}. This is due to its role as a framework that guides party identity and electoral strategies. Precise measurement of this concept is, therefore, key to understanding political actors' behavior and rhetoric.

Turning this theoretical concept into measurable scales involved developing several methods ranging from manual coding to advanced computational approaches. Most of the early methods were based on word dictionaries. These techniques use predefined lists of terms to detect ideological labels based on word frequencies \citep{laver_party_1992, laver2000-yq}. While they provided a foundation for later approaches, these early methods failed to capture context and rhetorical variation inherent in textual data. Wordfish and Wordscores are later-generation models that rely on statistical scaling to place texts on ideological dimensions \citep{laver_extracting_2003, slapin_scaling_2008}. While their measurements are more accurate, they also fail to incorporate contextual features.

Recent developments of machine learning further improved the techniques used to capture ideology in political texts. Supervised classification models rely on labeled datasets to identify ideological cues and inclinations \citep{hopkins_method_2010}. Pre-trained models address the issues of data scarcity and intensive annotation requirements while improving accuracy \citep{kim_new_2023, makhortykh_panning_2022}. Fine-tuning further improves their performance in domain-specific applications. Zero-shot and state-of-the-art generative models learn how to capture explicit and implicit trends in data without pre-labeled references \citep{buscemi_large_2024, krugmann_sentiment_2024}. This reduces resource requirements while improving accuracy.

The current availability of options when measuring ideology provides researches with ample opportunities to examine political phenomena. However, there is an increasing need to compare their effectiveness and identify the strengths and weaknesses of different models. Specifically, only a few systematic studies have compared fine-tuned and zero-shot transformer models and generative large language models (LLMs). Each approach entails a set of trade-offs related to accuracy, scalability, and resource requirements \citep{puri_zero-shot_2019,bucher_fine-tuned_2024}. Clarifying these trade-offs is essential to conducting research relying on ideology measurements.

This study relies on the replication materials provided by \citet{benoit_crowd-sourced_2016} to assess the performance of 12 machine learning models (and model variations) in detecting economic ideology in political text. The dataset contains 13,304 sentences from U.K. party manifestos along with manual annotations of economic ideology by experts and crowds. Building and expanding on \citet{lemens2024}, I compare the output of different models against human coding. I conduct the comparison on the granular sentence-level as well as the aggregate manifesto-level.

The results of the analysis reveal several key findings. Although not task-specific, generative LLMs are pre-trained on large amounts of data allowing for cross-task applications. Comparisons of sentence-level performance metrics and manifesto-level correlations show that they outperform other models. They consistently achieve the highest performance metrics and correlations with human coding. Fine-tuning involves training transformer models for domain-specific uses. The analysis shows that they can be a promising alternative to generative models. Although labor-intensive and their performance is lower, they offer the advantage of being open-source and requiring less computational resources. Zero-shot transformer models leverage general language understanding to perform tasks without prior training. Despite this flexibility, the results show that they struggle to identify economic ideological cues.

The contributions of this paper to the growing literature on computational methods are threefold. First, it provides one of the few systematic assessments of machine learning models' performance in detecting (economic) ideology. Second, it synthesizes the findings to highlight trade-offs. Generative LLMs achieve accurate and reliable performance but require intensive computational resources. Fine-tuned models achieve average performance and require pre-annotated training sets. However, they are accessible and resource-efficient. Zero-shot models offer flexibility but lack in accuracy. Third, this study provides best-practices. For instance, fine-tuning's performance is associated with the size of the training sets. Larger training sets (800 to 1000 texts) achieve better results. In addition, the performance of zero-shot models varies depending on the hypothesis template (or prompt).

By addressing these methodological questions, this study provides a framework to detect economic ideology in political text. It advances our understanding of this central concept in political science research. These findings highlight the need to adjust methodological approaches according to the research objectives. This research also provides a reference for researchers examining economic ideology depending on resource accessibility.

\section{Background}
\subsection{Dimensions of Economic Ideology \& Its Role in Shaping Political and Legislative Behavior}
Economic ideology can be defined according to several dimensions. However, its overarching themes revolve around the government's role in the economy. These themes span different areas such as regulation, redistribution, taxation, and government spending. Such aspects are key to examining political behavior and rhetoric, policy preferences, and the characteristics of institutions \citep{downs_economic_1957,lipset_political_1960}. Traditionally, economic ideology is organized around a left-right dimension. It ranges from state interventionism to free-market capitalism. This framework guides both individual-level and collective-level decision-making \citep{poole_congress_1997} as ``it does not simply reflect possible pathways through the political; it also plays a role in shaping it'' \citep[99]{stanley_2008}. Economic ideology is, therefore, one of the key elements that shape voters' and politicians' behavior.

The definition of economic ideology also encompasses a multidimensional aspect. It relates to several areas such as trade, labor regulations, financial markets, business environments, taxation, and social welfare. These dimensions interact and shape individual preferences on specific issues \citep{gabel_putting_2000, hooghe_does_2002}. However, this overarching nature led to the political competition literature frequently treating it as a single dimension \citep{benoit_party_2006,Marks2006-ma}. The multidimensional nature of economic ideology makes it adjustable to different country-level and institutional settings.

This concept also occupies a central role in the political sphere. \citet[13]{freeden_2006} argues that ``all expressions of political thought, irrespective of the various readings to which they may be subject, adopt the form of ideologies.'' Its underlying features affect policy-making, voting behavior, as well as coalition formation. For legislators, ideology affects their policy positions and partisan alignments \citep{kalt_capture_1984,proksch_position_2010}. For instance, the literature frequently demonstrates its impact on roll call voting patterns across different institutional settings. In this context, economic ideology is a key predictor of legislative outcomes \citep{hix_power_2005}. It also has a stabilizing influence in a time of extreme partisanship.

Another evidence of its overarching nature is the influence exerted by economic ideology on electoral behavior. Voters construct their electoral choices according to an interaction of key issue preferences, including on economic policy. Generally, positions on economic issues such as fiscal policy or corporate regulation align with party association \citep{iversen_asset_2001,huber_church-state_2011}. Similarly, incumbents usually face an assessment of their economic performance, although filtered by partisan perspectives, which drives electoral outcomes \citep{lewis-beck_economic_2000, duch_economic_2008}.

Economic ideology provides an opportunity for parties to exhibit points of differentiation and, hence, affect electoral competition. A number of comparative studies have demonstrated that, especially in multiparty systems, economic policy forms one of the foundational elements of party identity and strategy \citep{budge_mapping_2001, klingemann_mapping_2006}. Party manifestos also often follow shifts in public opinion and the salience of different economic issues \citep{adams_understanding_2004, ezrow_parties_2008}. As a result, this influence on partisan alignment and behavior underlines its role in structuring electoral competition.

Responses to different political and social phenomena are also guided by economic ideology. Globalization had different consequences across countries. In Western democracies, the alienation of blue-collar workers intensified the debates between protectionists and free-market proponents \citep{rodrik_has_1997, garrett_partisan_1998}. Economic downturns are also an example eliciting ideologically driven responses. The 2008 crisis polarized politicians and public opinion. It created a divide between those advocating for market solutions and those calling for more government regulation \citep{margalit_explaining_2013, pontusson_how_2012}. These patterns demonstrate the changing nature of economic ideology and its capacity to adapt to different realities.

\subsection{Measurement of Ideology in Political Texts}
As a central concept in political science research, ideology requires accurate measurements that reflect the realities of personal and collective preferences. \citet[98]{stanley_2008} underscores the role of language in shaping ideological beliefs: ``if ideas are individual interpretations, ideologies are interpretive frameworks that emerge as a result of the practice of putting ideas to work in language as concepts.'' Early efforts of operationalization relied heavily on manual coding of speeches and manifestos by experts. These approaches provided foundational structured insights into ideological trends among political actors. However, these approaches faced several limitations including their lack of scalability and issues of subjectivity by coders \citep{budge_mapping_2001, klingemann_mapping_2006}. Yet, these initial efforts provided the first frameworks to quantify ideology, paving the way for future systematic and replicable methods.

The significance of language is not in the syntax or the isolated terms within it; rather, it is the broader narrative and context it creates. This assertion summarizes the evolution of automated analyses of (political) text. Early methods relied on the mere presence of words to identify concepts. Starting with dictionary-based to the relatively more sophisticated bag-of-word or scaling approaches, early methods of analyses failed to incorporate context \citep{laver_extracting_2003, slapin_scaling_2008, hopkins_method_2010}. Later-generation methods attempted to address this concern to varying degrees of success \citep{blei_latent_2003}. Recent developments in machine learning allowed transformer models to contextualize terms within their surrounding lexical environment \citep{brown_language_2020}. This shift from the mere presence of words to capturing meaning and connotations allow text models to identify abstract patterns and implicit ideas. For legislative rhetoric, this means the ability to detect the tone and the context that construct these messages \citep{radford_language_2019, gentzkow_measuring_2019}.

Dictionary-based approaches improved upon human coding with systematic ideology detection. Using pre-selected word lists, researchers can detect ideological leanings through term frequencies. Previously done with manual annotations and labor-intensive processes, dictionary-based methods provide the opportunity to track ideological trends across documents. However, they fail to automatically account for linguistic evolution and contexts \citep{grimmer_text_2013, monroe_fightin_2008}.

Later-generation methods relied on latent variable models to position textual data on continuous ideological scales. These scaling techniques leverage distributions of word frequencies relative to reference texts to compute ideological scores. Wordscores, developed by \citet{laver_extracting_2003}, refers to pre-selected texts with known policy positions to assign ideological scores to words. Wordfish, based on the Poisson distribution and developed by \citet{slapin_scaling_2008}, detects patterns of word usage to classify texts interdependently. Although more accurate and scalable than dictionary-based approaches, scaling methods pose several limitations \citep{lowe_scaling_2011, lauderdale_measuring_2016}. Wordscores and Wordfish also lack contextual understanding along with issues with text selection.

Recent developments in machine learning and artificial intelligence addressed the challenges inherent to previous approaches. In addition to improved automation of textual analysis, transformer models leverage techniques such as word embeddings to account for contextual variations \citep{devlin_bert_2019, sanh_distilbert_2020, he_deberta_2021}. Through different attention mechanisms, they capture semantic relationships between words. This greatly improves upon the traditional methods based on word frequencies.

Supervised learning approaches entail detecting ideological labels or estimating positions based on pre-annotated data \citep{hopkins_method_2010, gentzkow_measuring_2019}. They offer ample opportunities to analyze large datasets but require extensive hand-coded training data. This requirement introduces issues related to potential biases and inconsistencies \citep{benoit_measuring_2019, teblunthuis_2024}. As a result, supervised methods come with a trade-off between accuracy and scalability for ideological measurement.

Unsupervised learning provides the benefit of analyzing textual data without requiring pre-annotated sets. Latent Dirichlet Allocation (LDA), for instance, identifies themes and patterns in text, which can be used to examine ideological dimensions \citep{blei_latent_2003, grimmer_bayesian_2010}. Unsupervised methods are particularly useful to conduct exploratory analyses and detect latent positions. The lack of human supervision, however, introduces issues related to capturing explicit ideological positions and mismatch between topics and ideological categories \citep{quinn_how_2010,roberts_structural_2014}.

The challenges presented by different ideological measurement techniques emphasize the importance of cross-validation against external benchmarks. Examples include comparing models' classification output with expert-coded data or voter surveys. This ensures reliability and interpretability in accurately measuring ideology \citep{benoit_treating_2009, gabel_putting_2000}. External cross-validation also addresses issues related to overfitting or misrepresentation \citep{caughey_dynamic_2015}. One of the solutions advanced by the literature entails adopting hybrid approaches. Human-in-the-loop frameworks integrate traditional methods with computational tools. For instance, combining expert coding with machine learning enhances performance and capturing contextual cues \citep{benoit_measuring_2019, denny_text_2018}. \citet[420]{Stromer_Galley_2023}, in their study classifying political campaign messages, advocate for combining human supervision with machine learning approaches. This preference over unsupervised or rule-based algorithms is due to ``the ability to develop categories that are grounded in existing theory and that have clear facial validity.'' Hybrid methods harness the computational power of machine learning while ensuring accuracy and efficiency.

\subsection{``All Quantitative Models of Language Are Wrong—But Some Are Useful''}
In their seminal paper, \citet[269]{grimmer_text_2013} lay out four principles for political text analysis, the first being ``all quantitative models of language are wrong—but some are useful.'' Machine learning offers researchers unprecedented pathways for computational methods and carries several trade-offs. Language models provide researchers with the opportunity to classify, summarize, and analyze large political textual data. As \citet[268]{grimmer_text_2013} note, ``automated content methods can make possible the previously impossible in political science: the systematic analysis of large-scale text collections without massive funding support.'' Scholars have rapidly integrated these methods for several tasks, ranging from classic uses such as policy area detection and sentiment analysis to more sophisticated concepts such as populism, nostalgia, and polarity \citep{mullerproksh2024, ballard_2022, moses_considerations_2021, wang_topic_2023, makhortykh_panning_2022, hajare_machine_2021, dehghani_political_2023, krugmann_sentiment_2024, Erhard_Hanke_Remer_Falenska_Heiberger_2025}. Several studies rely on machine learning to examine the dynamics of political discourse across various institutional settings \citep{erlich_multi-label_2022, van_atteveldt_studying_2019,mitrani_can_2022}. They also leverage these techniques to identify political content and ideological cues as they improve performance and adaptability. This extensive body of work showcases the promise of integrating machine learning to systematically analyze political text.

Recent progress in computational methods relies on pre-trained language models (PLMs). Their acquired knowledge and capabilities can be applied to political research through transfer learning and with limited labeled data \citep{kamal_learning_2023, kim_new_2023, buscemi_large_2024, evkoski_xai_2023, dehghani_political_2023, makhortykh_panning_2022, hajare_machine_2021, nguyen_how_2020}. They outperform classical models and further reduce data scarcity issues. PLMs facilitate the empirical examination of different political concepts in an automated way and even for resource-poor language and cross-lingual settings \citep{dehghani_political_2023, dyevre_text-mining_2021,le_mens_uncovering_2023}. They offer the chance to analyze semantic meaning and subtle ideological signals embedded in political discourse.

Applications of machine learning approaches to political phenomena include various sources such as party manifestos, legislative speeches, media articles, and social media content. This variety of sources enables cross-domain and cross-lingual comparative studies \citep{mullerproksh2024, osnabrugge_cross-domain_2021}. Overlapping with political discourse, researchers also extended these applications to analyses of legal documents \citep{dyevre_text-mining_2021}. Multilingual embeddings expanded the scope of political research to include non-English and cross-cultural contexts.

These advances come with several trade-offs when analyzing political text. The issues include data noise, political bias, contextual dependencies, and energy resources \citep{erlich_multi-label_2022, mitrani_can_2022, nguyen_how_2020, wilkerson_large-scale_2017}. These inherent challenges can be addressed by careful model selection and conducting robust validations along with domain knowledge integration. \citet[1]{baden_2021} highlight that the development of text analysis methods ``has prioritized technological over validity concerns, giving limited attention to the operationalization of social scientific measurements.'' Explainable AI, designed to provide clear explanations of its output, may also help in understanding model predictions when examining political concepts \citep{evkoski_xai_2023}.

Furthermore, the application of these methods to political research requires solid transparency and theoretical grounding \citep{pardo_Guerra_2022}. Researchers highlight the need for maintaining ethical standards and taking algorithmic bias into account \citep{grimmer_representational_2013, wilkerson_large-scale_2017,moses_considerations_2021}. In this context, balancing methodological innovation with empirical rigor enables reliable interpretations of political texts.

\subsection{Generative, Fine-tuned, \& Zero-Shot Machine Learning Models}
The key differences between generative, fine-tuned, and zero-shot models lie in how their knowledge is acquired and their adaptability to political textual analyses. Trained with self-supervised learning on a large amount of data, generative models capture statistical patterns in text. They reflect a broad understanding of linguistic syntax and semantics \citep{brown_language_2020, radford_language_2019}. Fine-tuning is a form of transfer learning whereby a pre-trained language model acquires task-specific knowledge. Training on pre-labeled example sets improves its accuracy and performance \citep{bucher_fine-tuned_2024,wang_selecting_2024}. In contrast, zero-shot models offer the advantage of not requiring labeled data. They execute new tasks and concepts on previously unseen examples based on a pre-training unlabeled vast amount of data. While they offer flexibility, they often lack in performance \citep{puri_zero-shot_2019,hu_leveraging_2024}.

In several research settings, issues of unavailability of annotated data or tasks requiring a broad understanding of language may arise. Generative models are particularly helpful in these cases. Their capabilities in inferring labels make them cost-effective as they do not require resource-intensive manual coding \citep{cai_monte_2024, puri_zero-shot_2019}. For political research, this is especially useful for studies examining political concepts requiring complex tasks such as event extraction or stance detection \citep{krugmann_sentiment_2024, li_learning_2024}. \citet{tornberg2024} finds that GPT-4 outperforms supervised models such as Naïve Bayes and human coders in detecting political affiliation. While these opportunities offer unprecedented pathways for computational political science, they carry limitations mainly linked to accessibility. While a few are open-source, most cutting-edge generative models are privately owned and charge for their services.

Fine-tuning involves combining large pre-trained language models with pre-labeled data. This integration greatly improves performance as they surpass their zero-shot counterparts in political text classification \citep{bucher_fine-tuned_2024,alizadeh_open-source_2024}. Through training on annotated data, they learn to distinguish and detect linguistic markers and domain-specific terminologies. Prior work shows that this improved predictive power makes them outperform zero-shot models in identifying political stances or sentiment \citep{bucher_fine-tuned_2024,wang_topic_2023,wang_selecting_2024}. Aligning the model's parameters with the requirement of a target set improves its capabilities in handling complex political content.

In contrast, zero-shot models do not require task-specific training as they leverage general language capabilities. These models are ideal for research settings where there is a lack of resources, coded data, and time \citep{hu_leveraging_2024, kuila_deciphering_2024}. Relying on its built-in parameters, researchers can use zero-shot learning to execute tasks via prompt engineering. Prior research points to improvement when combined with chain-of-thought or rationale-augmented procedures \citep{kuila_deciphering_2024,cai_monte_2024}. For political science research, these advantages provide novel possibilities to study emerging political events or innovative classification tasks. The trade-off lies in lower accuracy. Without recognizing domain-specific linguistic cues, zero-shot models generally suit time or resource sensitive scenarios.

Benchmarking against traditional transfer learning shows that large language models -such as GPT-3.5, GPT-4o, or Llama 2- consistently yield equal or better performance \citep{krugmann_sentiment_2024, alizadeh_open-source_2024}. Yet, other comparative studies suggest that properly fine-tuned models outperform generative LLMs. \citet{bucher_fine-tuned_2024} conduct a systematic comparison between generative and fine-tuned models across several classification tasks and text sources. Their results show that ``smaller BERT-style models significantly outperform generative AI models such as ChatGPT and Claude Opus (used in a zero-shot fashion) across all applications'' \citep[2]{bucher_fine-tuned_2024}. In addition, DeBERTa and BERT variants emerged as alternatives that improve factual recall and precision in political text classification \citep{he_deberta_2021, zhong_factual_2021}.

Several methods combine concepts of fine-tuning and zero-shot learning. Specifically, few-shot strategies entail providing models with a minimal number of examples to guide their output \citep{wang_selecting_2024, kuila_deciphering_2024}. \citet{hu_leveraging_2024} leverage existing domain expertise for political relation classification by feeding language models information from existing annotation codebooks. As a result, finding a balance between architectural variations and training methods becomes a function of individual research objectives. \citet[5]{Burnham_2024} notes that ``there is no best approach; rather, the goal is to optimize model performance, computational efficiency, and human workload.''

\section{Data and Methods}
This chapter builds and expands on the existing analysis by \citet{lemens2024}. Among other analyses, the authors conduct cross-model comparisons using the replication materials by \citet{benoit_crowd-sourced_2016}. The data include sentences from U.K. party manifestos spanning several elections. Sentences cover economic and social policy areas and were assigned ideological scores by human coders (experts and crowds). Thus, they constitute a suitable benchmark to compare models' performance against. \citet{lemens2024} process the manifesto data on the sentence-level and instruct large language models to detect ideological labels. They design specifically tailored prompts to query the models on each sentence's score. The output is based on a pre-defined ideological scale (0-100) following a left-right dimension. These scores are then averaged on the manifesto-level to obtain cross-comparable values.

The systematic comparison by \citet{lemens2024} includes several cutting-edge models including GPT variants, Llama, and Mixtral. The analysis relies on aggregate manifesto scores based on models' predictions. The findings suggest reliable performance metrics of large language models. These LLMs outperform crowd coders and consistently achieve 0.9+ correlation coefficients relative to expert coders.

\begin{figure}[ht]
	\centering
	\caption{Number of sentences by manifesto and election year.}
	\label{2fig:manif_sents}
	\includegraphics[width=0.8\linewidth]{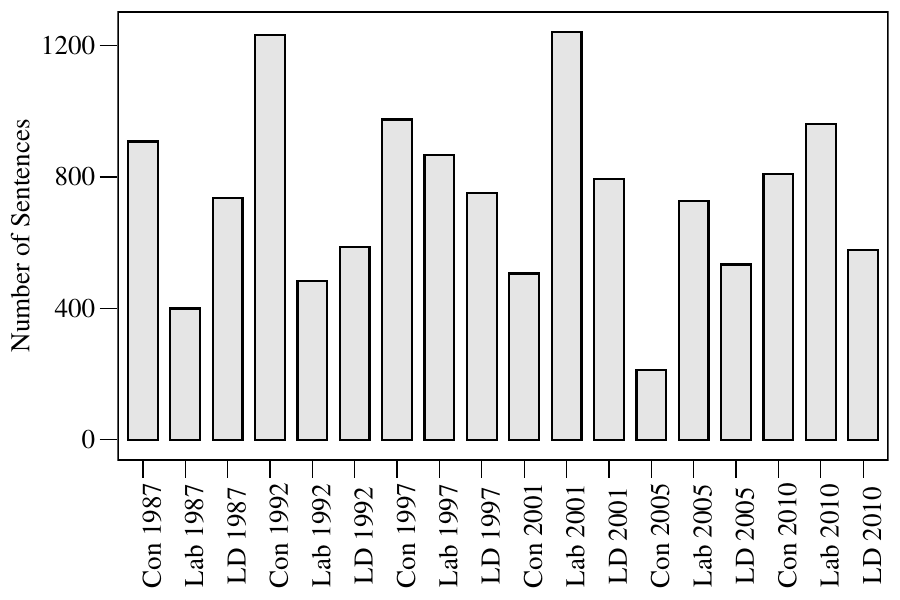}
\end{figure}

Building on the analysis by \citet{lemens2024}, I conduct a cross-examination of 12 machine learning models covering three categories: generative, fine-tuned, and zero-shot transformers. Table \ref{tab2:ai_models} provides an overview of these models. Several of these models can be used in both fine-tuned and zero-shot fashions. The analysis for DistilBERT includes both fine-tuned and zero-shot modes and the one for Gemini Flash includes both zero-shot and few-shot. For the few-shot iteration of Gemini, I provided the model one example sentence from each ideological label. For fine-tuning, I use a subset of 1000 pre-annotated manifesto sentences that are later excluded from the inference phase.

\begin{table}[htb!]
\centering
\caption{Overview of various models.}
\begin{threeparttable}
\resizebox{\textwidth}{!}{%
\begin{tabular}{|c c p{10cm}|}
\hline
\textbf{Model Type} & \textbf{Model Name} & \textbf{Description} \\
\hline
Generative & GPT-4o & GPT-4o is a state-of-the-art generative language model by OpenAI. Based on the transformer architecture, it is designed for complex reasoning and long-context understanding. \\ \hline
Generative & Gemini 1.5 Flash & Gemini 1.5 Flash was developed by Google DeepMind and integrates multimodal processing with high-speed generative abilities. It integrates language understanding and image recognition in a unified architecture.. \\ \hline
Fine-tuned & POLITICS\tnote{1} & POLITICS is a fine-tuned model based on RoBERTa and designed for political text analysis. It excels in tasks such as sentiment analysis and stance detection performed on political content. \\ \hline
Fine-tuned \& Zero-shot & DistilBERT\tnote{2} & DistilBERT is a distilled version of BERT offering a smaller and faster model while retaining BERT's performance. \\ \hline
Fine-tuned & RoBERTa Base\tnote{3} & RoBERTa Base is a robustly optimized BERT approach that is fine-tuned for improved performance in understanding and generating human language. \\ \hline
Zero-shot & DistilBART\tnote{4} & DistilBART is a distilled version of BART that is optimized for zero-shot classification tasks. It provides efficient text generation and understanding. \\ \hline
Zero-shot & DeBERTa\tnote{5} & DeBERTa is designed for zero-shot tasks. It offers improved language understanding via advanced attention mechanisms. \\ \hline
Zero-shot & RuBERT\tnote{6} & RuBERT is a Russian and English language adaptation of BERT. It was fine-tuned for natural language inference allowing for zero-shot classification. \\ \hline
Zero-shot & RoBERTa Large XNLI\tnote{7} & RoBERTa Large XNLI is fine-tuned on the XNLI dataset for cross-lingual understanding. \\ \hline
Zero-shot & Political DEBATE Large\tnote{8} & Political DEBATE Large is an NLI classifier trained for zero-shot and few-shot classification of political texts. It is optimally used in tasks such as stance detection and topic classification. \\ \hline
\end{tabular}%
}
\begin{tablenotes}
%\tiny
\scriptsize
\item[1] \url{https://huggingface.co/launch/POLITICS}
\item[2] \url{https://huggingface.co/distilbert/distilbert-base-uncased}
\item[3] \url{https://huggingface.co/FacebookAI/roberta-base}
\item[4] \url{https://huggingface.co/valhalla/distilbart-mnli-12-6}
\item[5] \url{https://huggingface.co/sileod/deberta-v3-base-tasksource-nli}
\item[6] \url{https://huggingface.co/cointegrated/rubert-base-cased-nli-threeway}
\item[7] \url{https://huggingface.co/joeddav/xlm-roberta-large-xnli}
\item[8] \url{https://huggingface.co/mlburnham/Political_DEBATE_large_v1.0}
\end{tablenotes}
\end{threeparttable}
\label{tab2:ai_models}
\end{table}

I use the manifesto data \citep{benoit_crowd-sourced_2016}, filtered for economic content only, and compare models' performance using two main approaches. First, I rely on sentence-level granular labels to compute performance metrics such as F1 scores and accuracy. Second, I aggregate the ideological predictions to the manifesto level and calculate correlation estimates relative to expert and crowd coders. The dataset comprises 13,304 sentences from the manifestos of three major U.K. parties (Conservative, Labour, Liberal Democrats). It spans six general elections from 1987 to 2010.

\begin{table}[htb!]
	\centering
	\caption{Contents of the prompts used to query different models.}
	\resizebox{\textwidth}{!}{%
		\begin{tabular}{|c p{12cm}|} % Add a vertical bar between columns for clarity
			\hline
			\textbf{Model} & \textbf{Prompt} \\ \hline
			GPT-4o & 
			``You will be provided with a text from a party manifesto. Where does this text stand on the left to right wing scale, in terms of economic policy? Provide your response as a label of ideology, either right-wing, left-wing, or neutral. To label the text, start by first identifying the parts of the text that are about economic policy. Then, determine the label based on these parts of text. You will only respond with a JSON object with the label. Do not provide explanations.'' \\ \hline
			Gemini 1.5 Flash & 
			``You will be provided with multiple texts from party manifestos. Where does each text stand on the left to right wing scale, in terms of economic policy? Provide your response as a list of labels of ideology, either right-wing, left-wing, or neutral. For each text, provide the label in the format \{text\_number: N, label: your\_label\_here\}. Do not provide explanations or any other text. Here are the texts:'' \\ \hline
			Gemini 1.5 Flash (Few-shot) & 
			``You will be provided with multiple texts from party manifestos. Where does each text stand on the left to right wing scale, in terms of economic policy? Provide your response as a list of labels of ideology, either right-wing, left-wing, or neutral. For each text, provide the label in the format \{text\_number: N, label: your\_label\_here\}. Do not provide explanations or any other text. Here are some examples: 1. ``We need to cut taxes and reduce regulations for businesses to stimulate economic growth.'' \{text\_number: 1, label: right-wing\} 2. ``Investing in public and affordable healthcare is crucial for our citizens.'' \{text\_number: 2, label: left-wing\} 3. ``In Britain today, living standards are higher than ever before in our history." \{text\_number: 3, label: neutral or procedural\} Here are the texts:'' \\ \hline
			All zero-shot models & 
			``Right-wing beliefs emphasize free-market capitalism, low taxes, free trade, deregulation, privatization, individualism, promoting the private sector, and limited government intervention. Left-wing beliefs emphasize government intervention, wealth redistribution, protectionism, progressive taxation, expanded welfare programs, and government regulation. Neutral refers to apolitical or factual content. The political economic ideology expressed in this statement is \{ \}.'' \\ \hline
		\end{tabular}%
	}
	\label{tab2:prompts}
\end{table}

The data from \citet{benoit_crowd-sourced_2016} includes multiple codings for each sentence by several expert and crowd coders (on a numerical scale). I follow a majority-based rule to assign a single categorical label to each sentence for experts and the same for crowd coders. For each type of coder, I group the sentences by identifier and extract the numeric value with the highest count (reflecting the classes left-wing, neutral, or right-wing). In the rare cases where this process results in a tie, I exclude the sentence to maintain consistency and avoid ambiguity in the analysis.

The first approach in the analysis entails comparing sentence-level performance metrics. At every iteration, the models predict a label for each manifesto sentence on the economy. The task is to choose among three ideological classes: left-wing, neutral, or right-wing. I then compare the predicted labels by each model with the expert and crowd codings resulting from the majority-based approach. This benchmarking against human coding allows me to compute several performance metrics such as F1 scores, accuracy, or recall.

At the aggregate level, manifesto-level ideological scores for human coding and models' predictions are based on different approaches. For human coding already included in the \citet{benoit_crowd-sourced_2016} data, I start by calculating the average of the ratings received by each sentence by expert and crowd coders separately. Then, I group the sentences by manifesto and calculate an overall average that reflects the ideological scores on a left-right dimension. These numerical scores by expert and crowd coders will later serve as a benchmark to assess models' performance.

\begin{figure}[ht]
	\centering
	\caption[Average economic ideological scores by manifesto based on crowd and expert codings.]{Average economic ideological score by manifesto based on crowd and expert codings. Negative values indicate a more left position. Positive values indicate a more right position.}
	\label{2fig:manif_pos}
	\includegraphics[width=0.8\linewidth]{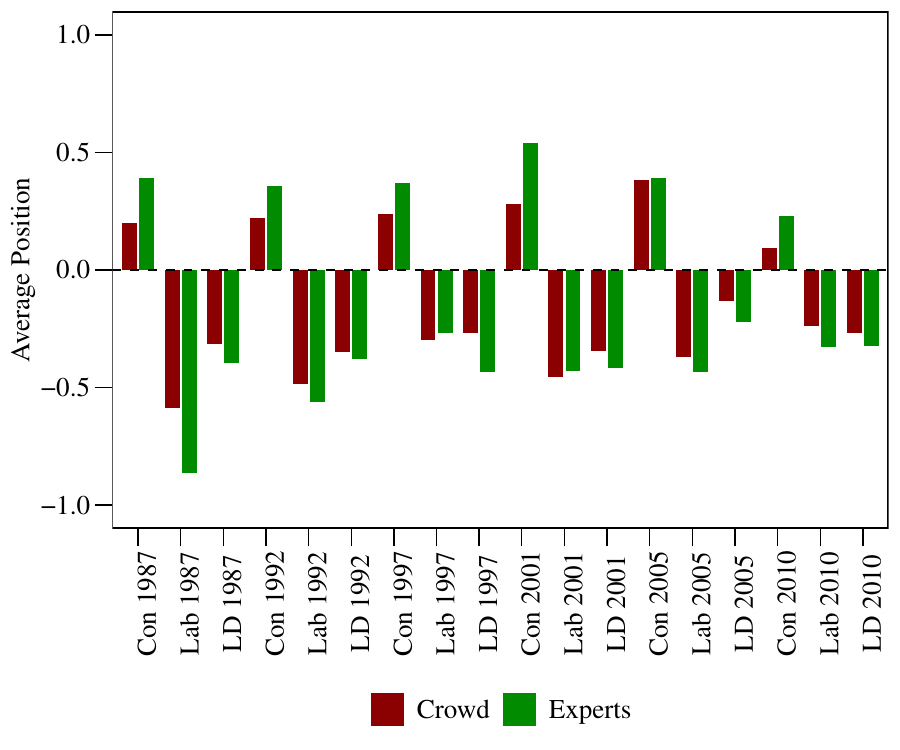}
\end{figure}

Figure \ref{2fig:manif_pos} showcases average economic ideological scores by manifesto based on crowd and expert codings. The values indicate a general alignment in trends between expert and crowd coders. However, they differ in magnitude with experts consistently assigning more extreme values on both sides of the aisle. A notable example is the 1987 manifesto of the Labour party where experts assigned a considerably more negative (left-wing) score compared to crowd coders. This is also the case for the 2001 Conservative manifesto where crowd coders appear to be more moderate than experts.

Operationalizing manifesto-level economic ideology scores based on models' output adopts a different approach. For each model and manifesto, I calculate the number of right-wing labels and the number of left-wing labels and use their logged ratio as an indicator of ideology. This follows established methods in the literature used to estimate policy positions \citep{lowe_scaling_2011} and sentiment in political texts \citep{proksch2019}. The resulting numerical values indicate the overall economic position of a given manifesto based on each model's predictions. I compare these values with the previously calculated average positions based on expert and crowd ratings using correlation coefficients. To facilitate cross-model interpretation, I standardize the economic ideology scores for each model to have a mean of 0 and a standard deviation of 1. I also do this for the manifesto ideological averages based on expert and crowd ratings.

\begin{singlespace}
	\begin{align*}
		\text{Ideology Score}_{p,t}=
		log\left(\frac{\text{Count}_{R,p,t}\:+\:0.5}{\text{Count}_{L,p,t}\:+\:0.5}\right)
	\end{align*}
\end{singlespace}

where:
\begin{itemize}
	\setlength{\itemsep}{0pt} % Space between items
	\setstretch{1}
    \item $\text{Count}_{R,p,t}$: The total number of right-leaning sentences in the manifesto of party $p$ in election year $t$.
	\item $\text{Count}_{L,p,t}$: The total number of left-leaning sentences in the manifesto of party $p$ in election year $t$.
\end{itemize}

These two approaches (sentence-level comparisons and manifesto-level correlations) ensure that the assessment covers both granular and aggregate measures of ideology. This expands on the analysis by \citet{lemens2024} as it only examines aggregate correlations. Sentence-level performance metrics reflect models' capabilities in accurately matching human coding. While aggregate correlations detect overarching economic ideology trends. As the results in the following section will demonstrate, these two performance assessments do not always align. In these cases, the criteria to choose a model depends on the specific research objectives.

\section{Results}
\subsection{Overview}
\begin{figure}[htb!]
	\centering
	\caption{Aggregate manifesto-level correlation coefficients by model - expert coding.}
	\label{2fig:overview_experts}
	\includegraphics[width=1\linewidth]{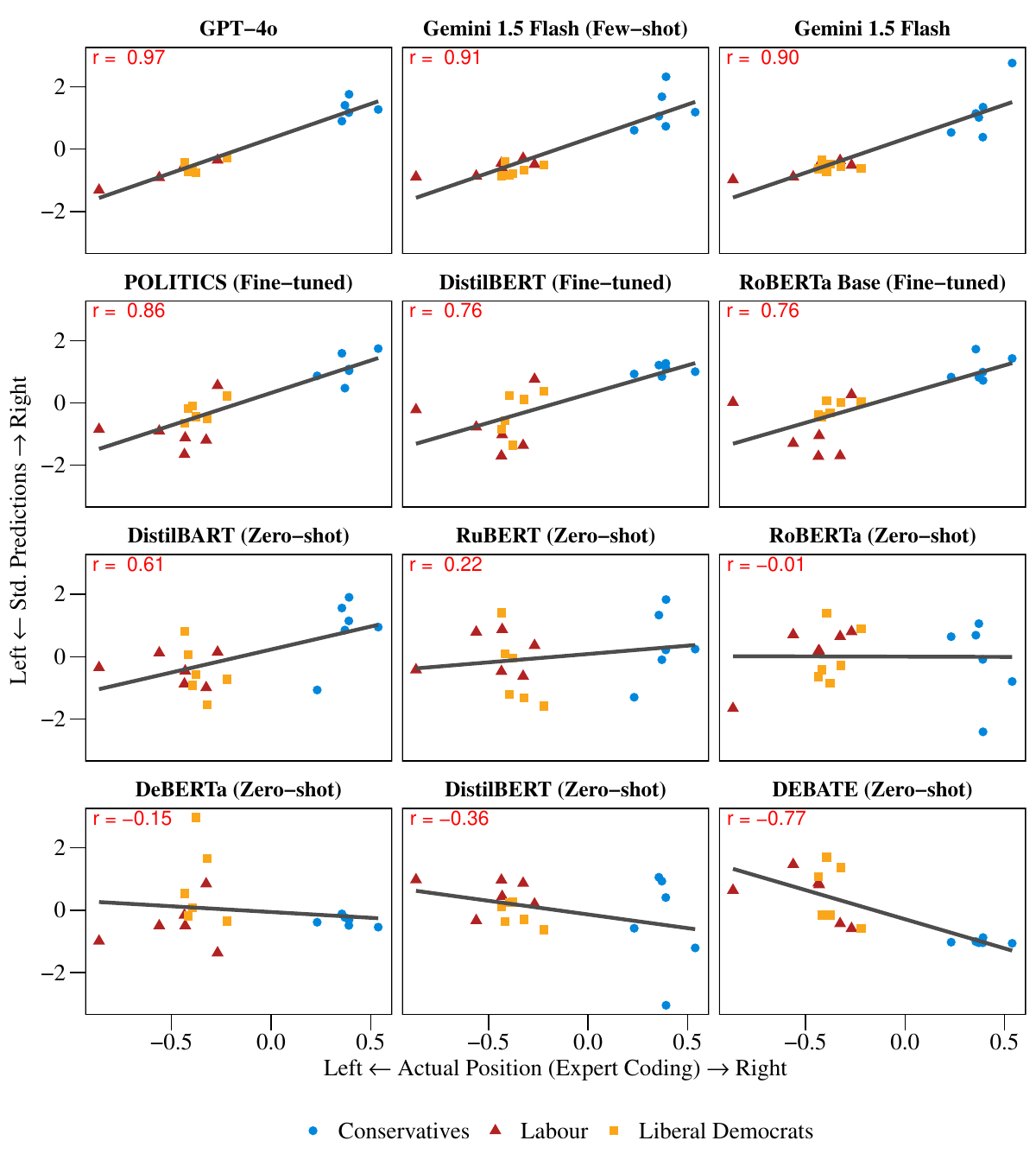}
\end{figure}

Before inspecting and validating each method, I present an overview of the main findings. Figures \ref{2fig:overview_experts} and \ref{2fig:overview_crowd} present cross-model correlation coefficients with the actual economic positions as identified by expert and crowd coders. I use the average ratings by experts as the main benchmark but also report comparisons with crowd coders.

Compared to expert ratings (Figure \ref{2fig:overview_experts}), generative models emerge as the best performing category in mirroring expert assessments at the manifesto level. GPT-4o achieves the highest correlation coefficient (0.97), followed by the few-shot iteration of Gemini 1.5 Flash (0.91) and Gemini 1.5 Flash (0.9). Although lower than their generative counterparts, fine-tuned models attain acceptable performance. The coefficients for fine-tuned POLITICS, DistilBERT, and RoBERTa Base are 0.86, 0.76, and 0.76, respectively. In contrast, zero-shot models reflect poor performance compared to expert ratings. While some zero-shot models return fair coefficients such as DisilBART (0.61), other models show a negative association with expert coding. For instance, the coefficients for DEBATE and the zero-shot iteration of DistilBERT are -0.77 and -0.36, respectively.

Figure \ref{2fig:overview_crowd} shows a similar pattern for correlations with crowd coding. Generative models prevail with correlations of 0.98 for GPT-4o, 0.94 for few-shot Gemini, and 0.88 for Gemini. Fine-tuned models retain their decent performance (POLITICS: 0.85; DistilBERT: 0.77; RoBERTa Base: 0.77). The performance of zero-shot either remains poor or worsens. For instance, the coefficients for RoBERTa and zero-shot DistilBERT go from -0.01 and -0.36 against expert ratings to -0.09 and -0.42 against crowd coding.

Correlation estimates with expert and crowd codings reveal that generative, followed by fine-tuned models, are suitable methods to detect economic ideology. Generative models benefit from their general language understanding and vast training data to capture ideological cues. Task-specific fine-tuned models were trained on a subset of manifesto sentences. This allows them to learn domain- and corpus-specific terminologies. Zero-shot models show that they struggle to detect economic ideology without pre-training for task-specific purposes. This initial overview provides two additional takeaways. First, most models, achieve higher correlation estimates with crowd coding compared to expert ratings. For instance, the coefficients for GPT-4o and Few-shot Gemini go from 0.97 and 0.91 against expert ratings to 0.98 and 0.94 against crowd coders. This is likely because crowd coders are more lenient whereas experts adhere to strict standards. Second, supplying generative models with a few examples improves their performance. The few-shot iteration of Gemini repeatedly outperforms its standard mode.

% latex table generated in R 4.3.1 by xtable 1.8-4 package
% Wed Dec 11 16:44:18 2024
\begin{table}[htbp]
\centering
\caption[Features with the highest keyness values in each target category by model.]{Keyness analysis by model. Table lists the 30 features with the highest keyness (chi-squared) values in each target category.} 
\label{tab2:keyness}
\begingroup\footnotesize
\begin{tabularx}{\textwidth}{|lXX|}
  \hline
 \textbf{Model} & \textbf{Left} & \textbf{Right} \\
  \hline
GPT-4o & poverty, education, women, training, tories, support, help, fund, access, commitment, work, public\_transport, teachers, system, including, making, community, powers, improve, aid, essential, nhs, poor, modern, liberal\_democrats, problems, rights, carers, propose, young\_people & competition, business, businesses, state, conservative\_government, regulations, enterprise, europe, best, lower\_taxes, whitehall, success, private\_sector, regulation, privatisation, wish, choice, savings, taxpayers, control, red\_tape, encouraging, unnecessary, saving, capital, bureaucracy, way, independence, shares, independent \\
\hline
  Gemini 1.5 Flash (Few-shot) & promote, environmental, care, life, support, liberal\_democrats, develop, women, alliance, improve, tackle, poverty, fairness, sports, training, food, promoting, public\_transport, access, environment, pollution, strengthen, nurses, banks, fair, supported, long-term, provision, carers, comprehensive & conservative\_government, state, private\_sector, conservative, lower\_taxes, regulations, choice, management, successful, europe, privatisation, whitehall, since, savings, directly, competition, accept, continue, bureaucratic, inflation, much, hard, strikes, business\_rates, regulation, control, allow, shares, believe, million \\
\hline 
  Gemini 1.5 Flash & poverty, environmental, tories, liberal\_democrats, care, women, alliance, tackle, develop, fairness, parliament, nhs, pollution, supported, communities, training, food, fair, modern, nurses, use, promote, comprehensive, young\_people, health, life, annual, fairly, improve, access & private\_sector, state, conservative\_government, regulations, savings, competition, lower\_taxes, inflation, successful, businesses, since, management, ownership, europe, enterprise, shares, become, public\_spending, conservative, hard, privatisation, keep, regulation, world, nation, monopoly, whitehall, choice, politicians, gain \\
  \hline
  POLITICS (Fine-tuned) & children, education, poverty, school, including, training, support, crime, help, parents, women, students, young\_people, programme, childcare, funding, work, improved, ensure, available, carers, nhs, aid, receive, tackle, schools, fund, payments, child, health & government, competition, business, enterprise, private\_sector, regulation, regulations, consumers, businesses, control, companies, privatisation, politicians, small\_businesses, freedom, red\_tape, conservative, market, bureaucracy, whitehall, local\_government, ownership, markets, private, state, governments, taxpayers, unions, trade\_union, monopoly \\
  \hline
  DistilBART (Zero-shot) & labour, labour's, liberal\_democrats, including, digital, labour\_government, poverty, progressive, earnings, commitment, immediately, currently, need, pensions, earning, package, welfare, help, one, decent, patient, promoting, leading, ensuring, supports, everyone, measures, nhs, address, incomes & conservatives, conservative\_government, must, schools, competition, companies, taxpayers, small\_businesses, local\_government, conservative, today, continue, individuals, teachers, oppose, heritage, privatisation, strong, crime, british, since, created, works, entrepreneurs, extended, countryside, mean, freedom, allowed, businesses \\ 
   \hline
\end{tabularx}
\endgroup
\end{table}

The Keyness analysis in Table \ref{tab2:keyness} lists the terms with the highest chi-squared values in each target category by model. It include the three generative models and the top performing fine-tuned and zero-shot models (POLITICS and DistilBART). Overall, the top terms for all model seem to reflect reliable economic ideological themes. Their predictions align with notable positions and issues associated with the left-right dimensions. DistilBART, however, appears to be heavily reliant on party labels.

For left-wing terms, the models consistently show terms aligning with social welfare and public services (poverty, nhs, education, public transport, training, aid, support, women, etc.). These features emphasize community support and public investment (training, funding, help). Gemini features terms related to the environment; an issue generally owned by left parties. Right-leaning terms with the highest values reflect a focus on creating a pro-business environment. These terms include private sector, red tape, management, business rates, ownership, among others. Another overarching theme is fiscal discipline with terms such as inflation, public spending, and taxpayers. These terms also convey anti-regulation and pro-competition stances (lower taxes, monopoly, capital, competition, etc.).

The terms in Table \ref{tab2:keyness} display a picture that follows conventional ideological divides. Overall, the models show common themes associated with each political family. Some distinctive features include GPT-4o's focus on a broad range of social and economic terms and Gemini's focus on environmental issues.

\subsection{Generative Models}
\label{sec:gen}

\begin{figure}[ht]
	\centering
	\caption{Performance metrics of different generative models.}
	\label{2fig:gen_f1_compare}
	\includegraphics[width=0.8\linewidth]{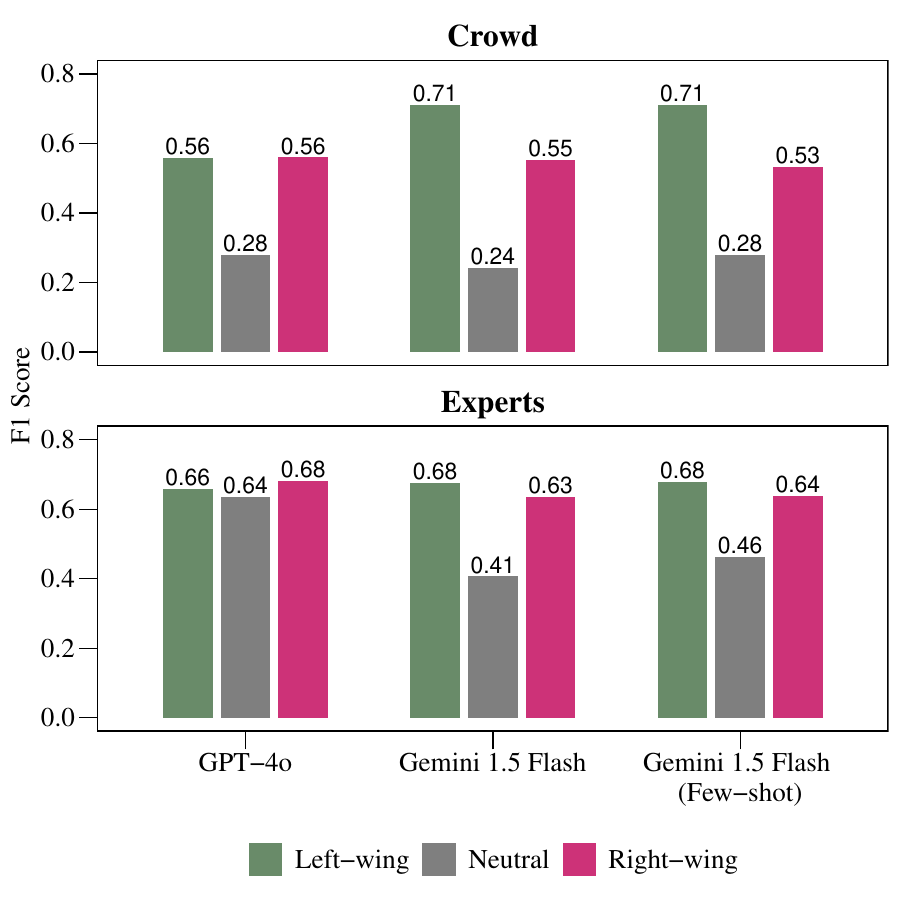}
\end{figure}

Assessments of sentence-level metrics for the generative category show reliable performance that varies across models. The sentences in Tables \ref{tab2:agreement_examples} and \ref{tab:disagreement_examples} show examples of agreement and disagreement between models and expert codings. In Table \ref{tab:disagreement_examples}, instances of disagreement are displayed in bold. Figure \ref{2fig:gen_f1_compare} and Table \ref{tab2:compare_gen} display F1 scores, a weighted mean of precision and recall. The analysis indicates several differences across models, across categories, and between expert and crowd coders' benchmarks. GPT-4o slightly outperforms other categories when compared to expert assessments. It achieves an F1 score of 0.66 for left-wing, 0.64 for neutral, and 0.68 for right-wing. This shows reliable capabilities in mirroring expert assessments at the granular level.

The standard mode of Gemini returns a slightly higher F1 score for left-wing (0.68) and a slightly lower score for right-wing (0.63). However, its performance considerably drops for the neutral category (0.41). This suggests that Gemini faces challenges with manifesto sentences that do not contain explicit ideological signals. The few-shot iteration of Gemini shows a similar pattern with for left-wing sentences (0.68) but shows a slight increase for neutral (0.46) and right-wing (0.64). Few-shot prompting, where the model is provided with a few examples, appears to slightly improve performance for these categories.

\begin{table}[htb!]
\centering
\caption{Agreement between models and expert coding on economic policy sentences.}
\label{tab2:agreement_examples}
\fontsize{10}{10}\selectfont
\setlength{\tabcolsep}{3pt} % Adjust column separation
\renewcommand{\arraystretch}{1.2} % Adjust row height
\begin{tabularx}{\textwidth}{|X|c|c|c|c|}
\hline
\textbf{Text} & \textbf{Experts} & \textbf{GPT-4o} & \textbf{Gemini FS} & \textbf{Gemini} \\
\hline
Our people are seeing their services cut, their waiting lists lengthened, and more and more needs going unmet. & Left-Wing & Left-Wing & Left-Wing & Left-Wing \\
\hline
We would restore grants for poor students and access to benefits for all during the summer holidays, and raise the salary threshold at which student loans are repaid, in the first instance from £10,000 to £13,000 per year. & Left-Wing & Left-Wing & Left-Wing & Left-Wing \\
\hline
To help achieve this aim we will double arts funding within the lifetime of one Parliament. & Left-Wing & Left-Wing & Left-Wing & Left-Wing \\
\hline
We will immediately increase the science budget to 0.35\% of GDP, and raise it steadily thereafter. & Left-Wing & Left-Wing & Left-Wing & Left-Wing \\
\hline
We must as a priority tackle the immediate tragedy and waste of unemployment. & Left-Wing & Left-Wing & Left-Wing & Left-Wing \\
\hline
Competition and open markets are by far the best guarantee of wealth creation. & Right-Wing & Right-Wing & Right-Wing & Right-Wing \\
\hline
Lower taxation, by increasing take-home pay without adding to industry's costs, improves competitiveness and helps with jobs. & Right-Wing & Right-Wing & Right-Wing & Right-Wing \\
\hline
Remotivating individuals and providing the right conditions for business are the only ways to make lasting change. & Right-Wing & Right-Wing & Right-Wing & Right-Wing \\
\hline
We will devolve as many powers as possible to schools and give them more control over their budgets. & Right-Wing & Right-Wing & Right-Wing & Right-Wing \\
\hline
What would postpone recovery, and turn this promise of growth into the certainty of hard times, is the election of our opponents whose policies would mean higher taxes, higher inflation, higher interest rates, more bureaucratic regulation and more strikes. & Right-Wing & Right-Wing & Right-Wing & Right-Wing \\
\hline
We will insist that the first call on income from the sale of mental health hospitals is the provision of better accommodation and services in the community for mental health users and people with learning disabilities. & Neutral & Neutral & Neutral & Neutral \\
\hline
Governments should learn to listen to the people to whom they are accountable. & Neutral & Neutral & Neutral & Neutral \\
\hline
The Scottish Parliament will have a vital role in building the competitive strength of the Scottish economy. & Neutral & Neutral & Neutral & Neutral \\
\hline
We will encourage our great national museums and galleries to make the national treasures which they house more widely accessible. & Neutral & Neutral & Neutral & Neutral \\
\hline
We will: Secure stable prices and low interest rates. & Neutral & Neutral & Neutral & Neutral \\
\hline
\end{tabularx}
\end{table}

\begin{table}[htb!]
\centering
\caption{Disagreement between models and expert coding on economic policy sentences.}
\label{tab:disagreement_examples}
\fontsize{10}{10}\selectfont
\setlength{\tabcolsep}{3pt} % Adjust column separation
\renewcommand{\arraystretch}{1.2} % Adjust row height
\begin{tabularx}{\textwidth}{|X|c|c|c|c|}
\hline
\textbf{Text} & \textbf{Experts} & \textbf{GPT-4o} & \textbf{Gemini FS} & \textbf{Gemini} \\
\hline
World Class Health and Public Services. NHS FUNDING PLEDGE: Continue, year by year, to increase the real resources committed to the NHS, so NHS spending will continue to share in a growing economy. & Left-Wing & Left-Wing & Left-Wing & \textbf{Right-Wing} \\
\hline
The choice for 2010. The Tories are the party of pensioner poverty. & Left-Wing & Left-Wing & Left-Wing & \textbf{Neutral} \\
\hline
Child Benefit will remain the cornerstone of our policy for all families with children. & Left-Wing & Left-Wing & Left-Wing & \textbf{Neutral} \\
\hline
We will continue that programme of modernisation. & Left-Wing & \textbf{Neutral} & \textbf{Neutral} & \textbf{Right-Wing} \\
\hline
Small firms will be assisted with a new investment scheme, combining a cash-limited fund for new investments with tax incentives tailored to their special needs. & Neutral & \textbf{Right-Wing} & \textbf{Left-Wing} & \textbf{Left-Wing} \\
\hline
We will give patients more choice. & Right-Wing & Right-Wing & \textbf{Neutral} & \textbf{Left-Wing} \\
\hline
We will extend charitable status to all schools without affecting total Council funding and maintain the VAT exemption on school fees. & Right-Wing & Right-Wing & \textbf{Left-Wing} & \textbf{Neutral} \\
\hline
Cutting subsidies. & Right-Wing & Right-Wing & \textbf{Left-Wing} & \textbf{Left-Wing} \\
\hline
And where there is a need for government intervention, for example in tackling pollution, our promise is to harness market forces with incentives for sustainable development, so that good businesses pay less tax. & Right-Wing & Right-Wing & \textbf{Left-Wing} & \textbf{Left-Wing} \\
\hline
The key priorities are: to deliver more choice and lower prices through liberalisation of financial services and utilities. & Right-Wing & Right-Wing & \textbf{Left-Wing} & \textbf{Left-Wing} \\
\hline
We will reform the system to increase choice, flexibility and support for working families. & Right-Wing & Right-Wing & \textbf{Neutral} & \textbf{Neutral} \\
\hline
We believe that franchising provides the best way of achieving that. & Right-Wing & Right-Wing & Right-Wing & \textbf{Neutral} \\
\hline
The operating side of BR will continue to provide passenger services until they are franchised out to the private sector. & Right-Wing & Right-Wing & Right-Wing & \textbf{Neutral} \\
\hline
Successful train operating companies will have their franchises extended to allow companies to invest in improved stations, car parks, facilities and rolling stock. & Right-Wing & Right-Wing & Right-Wing & \textbf{Neutral} \\
\hline
We will: Seek further reform of the Common Agricultural Policy (CAP). & Right-Wing & Right-Wing & Right-Wing & \textbf{Neutral} \\
\hline
\end{tabularx}
\end{table}

While I use expert assessments as the main benchmark, I also report performance metrics relative to crowd coders. The top panel of Figure \ref{2fig:gen_f1_compare} paints a different picture. Both iterations of Gemini slightly increase their performance in left-wing category compared to crowd ratings. However, across the three models, performance seems to drop. For the neutral label, F1 scores decrease significantly, at times more than halving. These range from 0.24 for standard Gemini to 0.28 for few-shot Gemini and GPT-4o. The right-wing label shows a similar drop with F1 scores of 0.56 for GPT-4o, 0.55 for few-shot Gemini, and 0.53 for standard Gemini. At the sentence level, the three generative models seem to align more with expert assessments. This difference is potentially due to the more consistent and standardized nature of expert codings.

Overall, sentence-level performance metrics show that generative models yield reliable results when predicting economic ideology. From a general view, GPT-4o achieves slightly better performance. However, running it required using the paid service by OpenAI. At the time of writing this chapter, Google offers the Gemini API for free with generous rate limits. Both iterations of Gemini achieve results comparable to those by GPT-4o, with a decrease in the neutral category. This makes it a strong resource-sensitive alternative.

\begin{figure}[htb!]
	\centering
	\caption{Aggregate manifesto-level correlation coefficients of generative models - expert coding.}
	\label{2fig:gen_experts_corr}
	\includegraphics[width=1\linewidth]{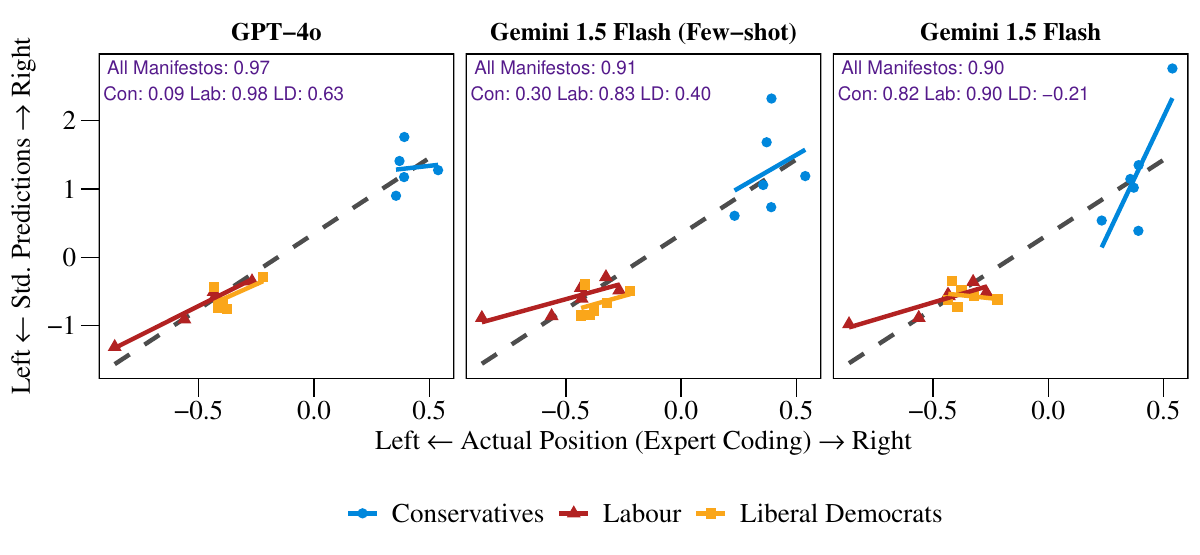}
\end{figure}

Figures \ref{2fig:gen_experts_corr} and \ref{2fig:overview_crowd} display manifesto-level correlation coefficients of generative models. Focusing on performance against expert ratings (Figure \ref{2fig:gen_experts_corr}), GPT-4o yields the highest overall correlation estimate (0.97). At the party level, GPT-4o performs exceptionally well with Labour manifestos (0.98) and moderately with Liberal Democrats manifestos (0.63). However, it struggles with mirroring expert assessments of Conservative manifestos (0.09). The average manifesto scores by expert and crowd coders in Figure \ref{2fig:manif_pos} show that Labour and Liberal Democrats manifestos are generally more polarized than Conservative ones. This could explain why GPT-4o performs well with the former as they likely contain more explicit ideological signals.

The standard mode of Gemini also shows reliable correlations with expert annotations. The overall estimate is 0.9. Within parties, Gemini returns a coefficient of 0.82 for Conservative manifestos (a substantial improvement on GPT-4o) and 0.9 for Labour parties. However, it encounters difficulties with matching expert ratings of Liberal Democrat manifestos with a negative correlation (-0.21). Few-shot prompting improves the overall performance of Gemini (0.91) and makes it better at detecting the centrist positions of Liberal Democrats (0.4). However, it slightly reduces performance with Labour manifestos (0.83) and significantly with Conservative ones (0.3).

In contrast to sentence-level performance metrics, GPT-4o and few-shot Gemini return higher correlation estimates with crowd coders than experts (0.98 and 0.94) and a slightly lower coefficient for standard Gemini (0.88). Figure \ref{2fig:gen_crowd_corr} also shows notable improvements in within-party performance for GPT-4o: 0.84 for Conservative manifestos, 0.95 for Labour, and 0.95 for Liberal Democrats. Both iterations of Gemini appear to face difficulties in identifying economic ideology labels within Liberal Democrats manifestos. This suggests cross-model challenges with identifying centrist positions lacking extreme ideological cues.
\subsection{Fine-tuned Models}
Fine-tuning involves training language models on task-specific terminologies and syntax. I assess the capabilities of three models available on the HuggingFace\footnote{www.huggingface.co} platform in detecting economic ideology. I extract a subset of 1000 manifesto sentences from the \citet{benoit_crowd-sourced_2016} replication data and use their corresponding expert codings in the fine-tuning process. This training sample was later excluded from the inference process.

\begin{figure}[ht]
	\centering
	\caption{Performance metrics of different fine-tuned models.}
	\label{2fig:finetune_f1_compare}
	\includegraphics[width=0.8\linewidth]{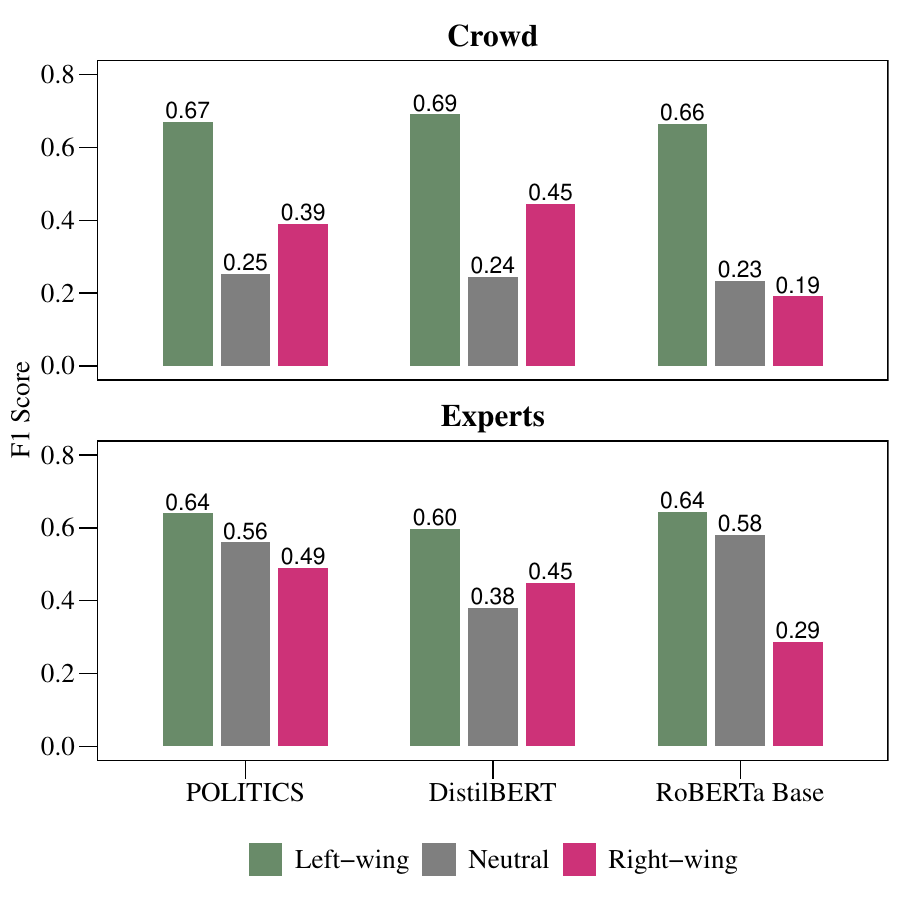}
\end{figure}

Figure \ref{2fig:finetune_f1_compare} and Table \ref{tab2:compare_finetuned} present the performance metrics of three fine-tuned models at the sentence level. These models demonstrate a consistent ability to correctly identify left-leaning sentences. This applies when compared to both expert and crowd coders. Compared to expert annotations, POLITICS, DistilBERT, and RoBERTa Base achieve F1 scores of 0.64, 0.6, and 0.64, respectively. This performance in the left-wing category improves when compared to crowd coders with F1 scores of 0.67, 0.69, and 0.66. These metrics reflect a general capability to correctly detect left-leaning economic ideology on both benchmarks.

Fine-tuned models exhibit lower performance in dealing with neutral language. Comparisons with expert assessments return F1 scores of 0.56, 0.38, and 0.58 for POLITICS, DistilBERT, and RoBERTa Base, respectively. Comparisons with crowd coders show a drastic decrease in performance with F1 scores of 0.25, 0.24, and 0.23. This mirrors the trend observed with generative models (Figure \ref{2fig:gen_f1_compare}) and suggests cross-model challenges with capturing neutrality compared to the less consistent crowd coding.

For right-wing sentences, the fine-tuned models face challenges with detecting economic ideology across both coding sources. Against expert ratings, the models return lower F1 scores than the left-wing label (POLITICS: 0.49; DistilBERT: 0.45; RoBERTa Base: 0.29). Against crowd annotations, F1 scores decrease to 0.39, 0.45, 0.19. This reflects challenges faced by fine-tuned models to capture right-leaning signals.

\begin{figure}[ht]
	\centering
	\caption{Aggregate manifesto-level correlation coefficients of fine-tuned models - expert coding.}
	\label{2fig:finetune_experts_corr}
	\includegraphics[width=1\linewidth]{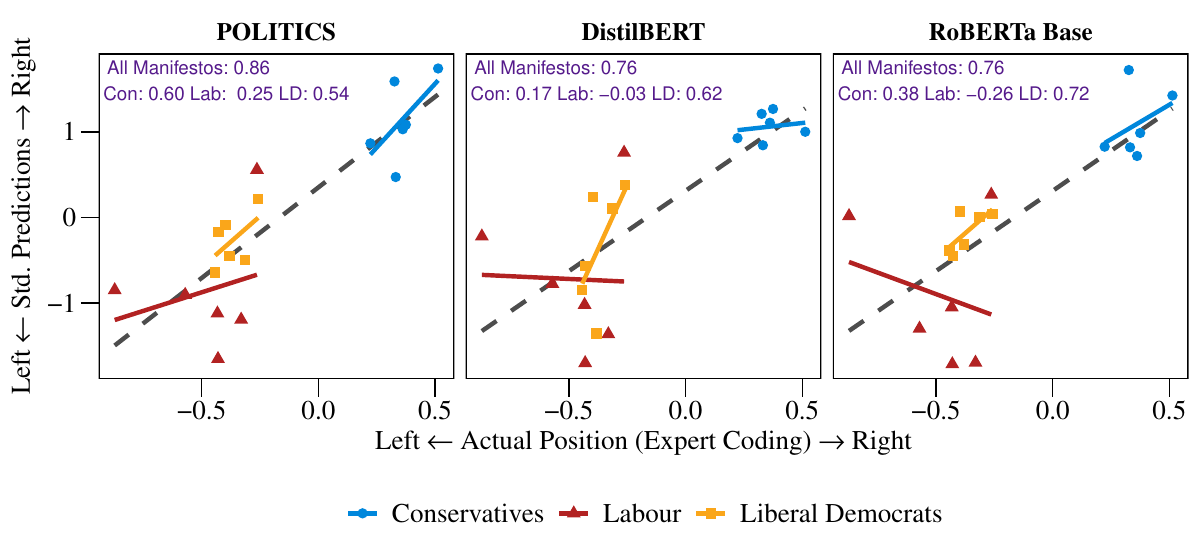}
\end{figure}

Aggregate-level performance indicators demonstrate the relative reliability of fine-tuned models in recognizing overarching ideological trends. Figure \ref{2fig:finetune_experts_corr} displays the estimates of correlation with expert judgments. All three models appear to closely follow expert ratings of economic ideology. POLITICS leads with an overall coefficient of 0.86, followed equally by DistilBERT and DeBERTa Base (0.76). However, the analysis shows varying patterns of within-party performance.

All three fine-tuned models return low, and at times negative, correlations with expert scores for Labour manifestos. This observation is particularly pronounced for RoBERTa Base with a coefficient of -0.26. In contrast, its ability to correctly identify overarching ideology in Liberal Democrats and Conservative manifestos is significantly better (0.72 and 0.38). POLITICS retains its top position among fine-tuned models with coefficients of 0.6 for Conservative manifestos, 0.25 for Labour, and 0.54 for Liberal Democrats. These results are counterintuitive as they contrast with the sentence-level metrics in Figure \ref{2fig:finetune_f1_compare}. The fine-tuned models returned the highest F1 scores in the ``left-wing'' category but struggle to output a holistic measure of Labour manifestos. This shows that granular level validation does not always align with aggregate correlations.

The estimates of correlation with crowd measurements in \ref{2fig:finetune_crowd_corr} follow a similar pattern. While they retained the overall acceptable coefficients, within-party performance of fine-tuned models against crowd codings significantly declines. For instance, POLITICS' coefficient for Conservative manifestos drops from 0.6 in Figure \ref{2fig:finetune_experts_corr} to 0.18 in \ref{2fig:finetune_crowd_corr}. RoBERTa Base's performance for Conservative manifestos goes from 0.38 to -0.1. The coefficients for Liberal Democrats also show equivalent dynamics across models.

As mentioned in section \ref{sec:gen}, the difference in performance indicators between the two benchmarks can be due to the variability characterizing crowd coding. There is an inherent subjectivity in individual interpretations of political content. The experts employed for the \citet[283]{benoit_crowd-sourced_2016} study are all academic researchers. Their coding process is arguably more methodological and objective. Therefore, crowd coders are more likely to be affected by this inherent subjectivity. Language models are trained on large datasets that include academic articles and other authoritative works. They also learn to recognize statistical and semantic patterns. These considerations might explain why they are more likely to align with expert coders' objectivity.

\subsection*{Fine-tuning and Training Data Size}
Fine-tuned models are heavily dependent on both the quality and the size of training sets. Output quality is as reliable as the input's. The issue of the size of training data is prevalent in the computational literature. However, the recommendations vary depending on model type and different requirements. For instance, \citet{laurer_less_2024} find that ``BERT-NLI performs better with very little training data ($\leq$1,000), while BERT-base is better when more data are available.'' As the best performing fine-tuned model, I use POLITICS to assess its capabilities in detecting economic ideology at different sizes of the training set.

\begin{figure}[htb!]
	\centering
	\caption{Performance metrics of fine-tuned POLITICS at different sizes of the training set.}
	\label{2fig:finetune_optimize}
	\includegraphics[width=0.8\linewidth]{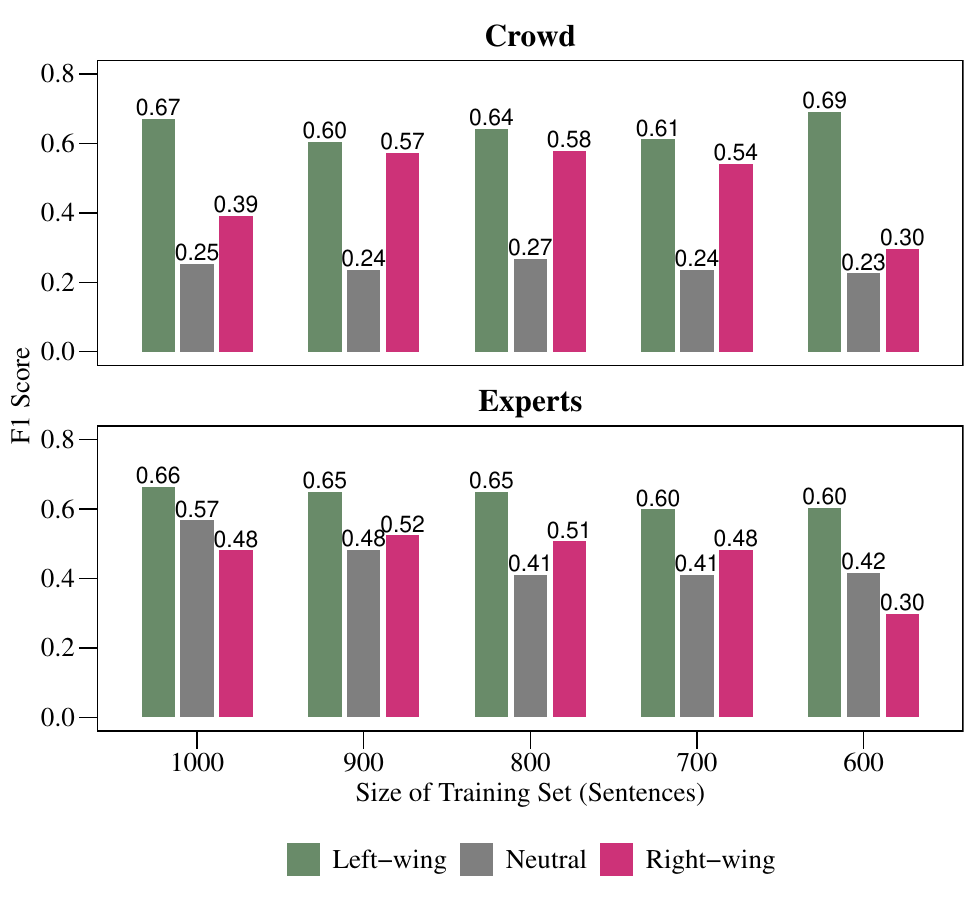}
\end{figure}

The initial fine-tuning was executed using a random set of 1000 manifesto sentences and their corresponding expert annotations. As Figure \ref{2fig:finetune_optimize} and Table \ref{tab2:finetuned_optimize} illustrate, I assess POLITICS' performance with additional training sets of 900, 800, 700, and 600 manifesto sentences. For left-leaning sentences and with experts as a benchmark, the model maintains relative consistency across sizes, although performance marginally declines with smaller sets. Below 800 sentences, F1 scores drop from $\sim$0.65 to 0.6 for the left-wing content. While the left-wing category seems less sensitive to training data size, the marginal improvement with larger sets can make a considerable difference.

Detecting neutrality appears to be more susceptible to changes in training data. The results show a positive association between dataset size and performance for this category. At 1000 sentences, the F1 score for the neutral category is 0.57. This decreases to 0.48 at 900 and drops further to $\sim$0.41 at 800 sentences and below. This suggests that fine-tuned models require more data for factual content or texts without clear ideological cues.

Compared to fine-tuning with 1000 sentences, F1 scores for the right-wing category record a slight improvement from 0.48 to 0.52/0.51 at 900/800 sentences. With the 700-sentence set, the F1 score goes back to 0.48 but significantly declines to 0.3 at 600 sentences. These values imply that substantially reducing training data negatively affects the right-wing label predictions.

\begin{figure}[htb!]
	\centering
	\caption{Aggregate manifesto-level correlation coefficients of fine-tuned POLITICS at different sizes of the training set - expert coding.}
	\label{2fig:optimize_experts_corr}
	\includegraphics[width=0.8\linewidth]{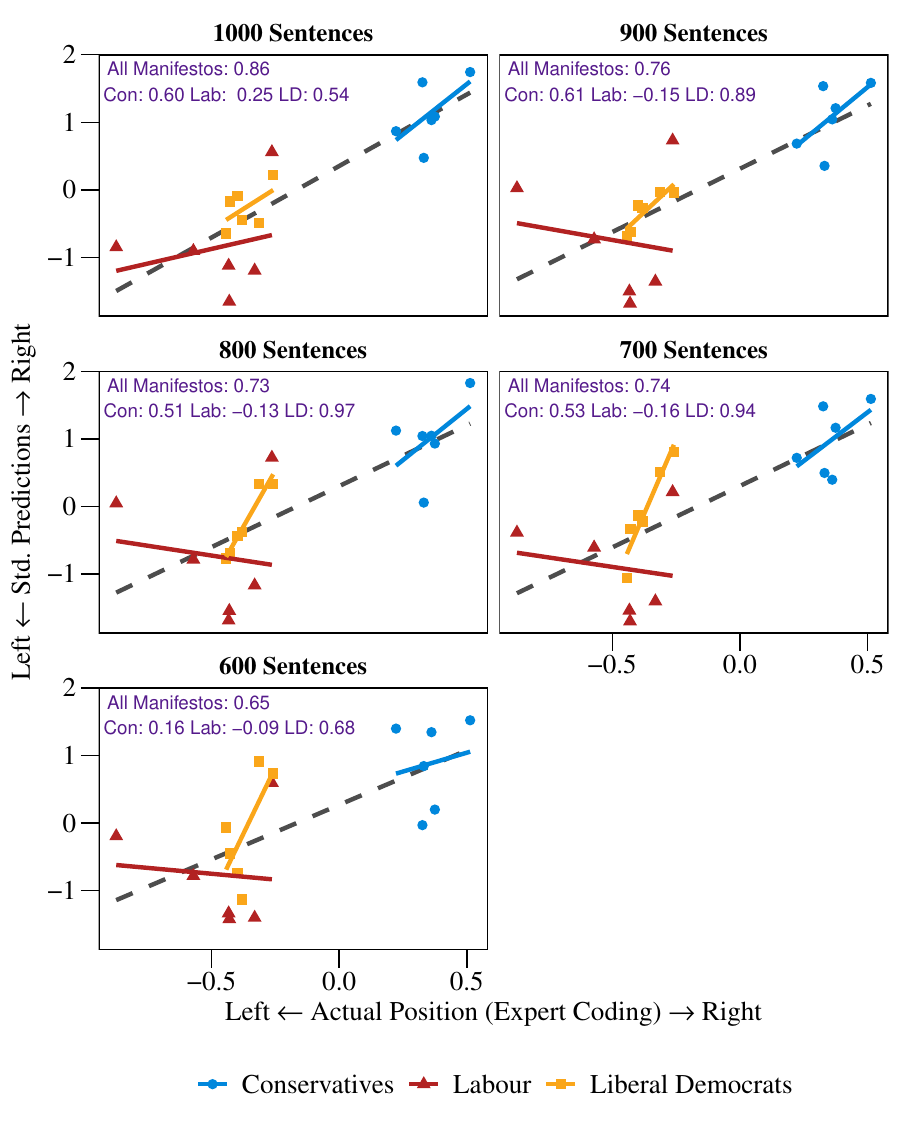}
\end{figure}

Holistic correlation estimates also reveal a positive association between training size and performance. This further demonstrates POLITICS' ability to capture aggregate ideological dynamics with larger sets. Figure \ref{2fig:optimize_experts_corr} (expert coding) shows that, for the overall scores, the best performance is achieved with a training set of 1000 manifesto sentences (0.86). Smaller sets return lower coefficients of 0.74, 0.73, and 0.76 between 700 and 900 sentences. Similar to sentence-level F1 scores, the training set of 600 sentences results in a significant drop (0.6). These trends confirm that larger training sets lead to better detection of overarching patterns of economic ideology.

Party-specific scores also reflect this positive relationship. The correlations for Conservative manifestos go from 0.16 at 600 sentences to 0.6 at 1000 sentences. Labour manifesto coefficients improve from -0.09 at 600 sentences to 0.25 at 1000 sentences. Liberal Democrats manifestos make the exception with significant (and varying) increases in performance at 900 sentences and below. A potential explanation could be linked to Liberal Democrats' centrist ideology. Smaller training sets arguably allow the model to identify centrist patterns which diminishes as more ideological content is introduced with larger sets.

\subsection*{Fine-tuning and Transfer Learning}

Although lower than their generative counterparts, the analysis shows that fine-tuned models achieve acceptable performance while being accessible and resource-sensitive. However, the previous performance metrics rely on fine-tuning and inference processes that are both based on the same type of data. A true assessment of accuracy would be predicting other types of data with the model fine-tuned on manifesto sentences.

\begin{figure}[htb!]
	\centering
	\caption{Performance of POLITICS fine-tuned on manifesto data when classifying parliamentary speeches.}
	\label{2fig:finetune_transfer}
	\includegraphics[width=0.8\linewidth]{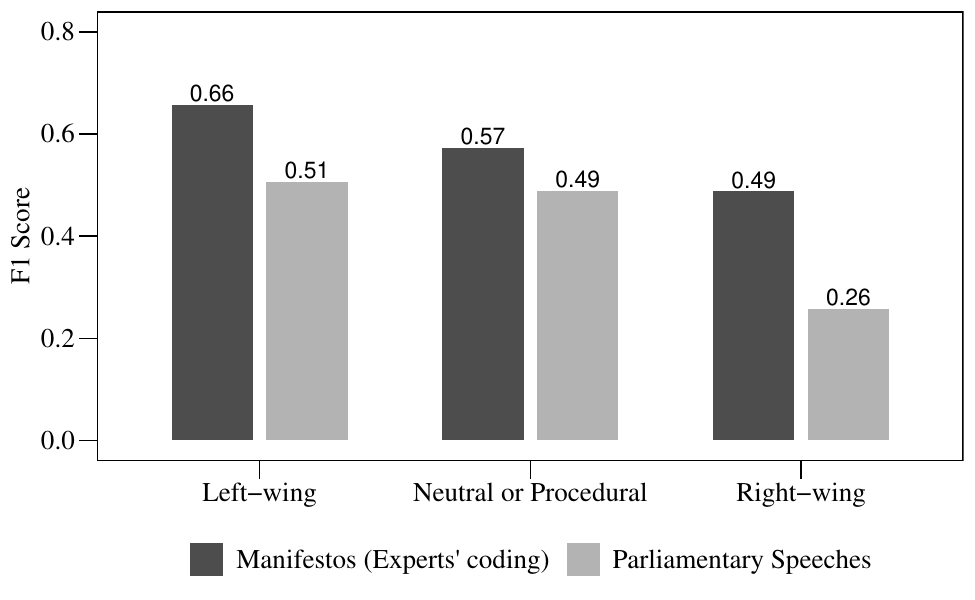}
\end{figure}

The results in Figure \ref{2fig:finetune_transfer} indicate that the model's performance is indeed affected when applied to another domain. I manually annotate the economic ideology of 900 sentences extracted from speeches in the U.K. House of commons. Using the POLITICS model fine-tuned on manifesto sentences, I predict the positions of the speech sentences and compute granular performance metrics. The F1 scores for parliamentary speeches for left-wing, neutral, and right-wing are 0.51, 0.49, and 0.26, respectively. The results for the same categories using manifesto sentences were 0.66, 0.57, and 0.49. This drop in performance implies that the capabilities of fine-tuning in detecting economic ideology are closely associated with the characteristics of the training set.

These findings reveal a severe limitation of fine-tuned models. Optimizing their accuracy requires providing the same type of annotated data as the ones used for predictions. As a result, while accessible and resource-sensitive, fine-tuning requires time-consuming manual annotations. This can be challenging especially when dealing with several types of political texts. For instance, in Chapter \ref{chap:ideology}, I analyze Tweets and parliamentary speeches from three countries. Identifying economic content with fine-tuned DistilBERT required annotating six separate cross-country and cross-platform training sets. These challenges highlight the advantage of generative and zero-shot models which do not require pre-labeled data.
\subsection{Zero-shot Models}
\begin{figure}[htb!]
	\centering
	\caption{Performance metrics of different zero-shot models.}
	\label{2fig:zeroshot_f1_compare}
	\includegraphics[width=0.8\linewidth]{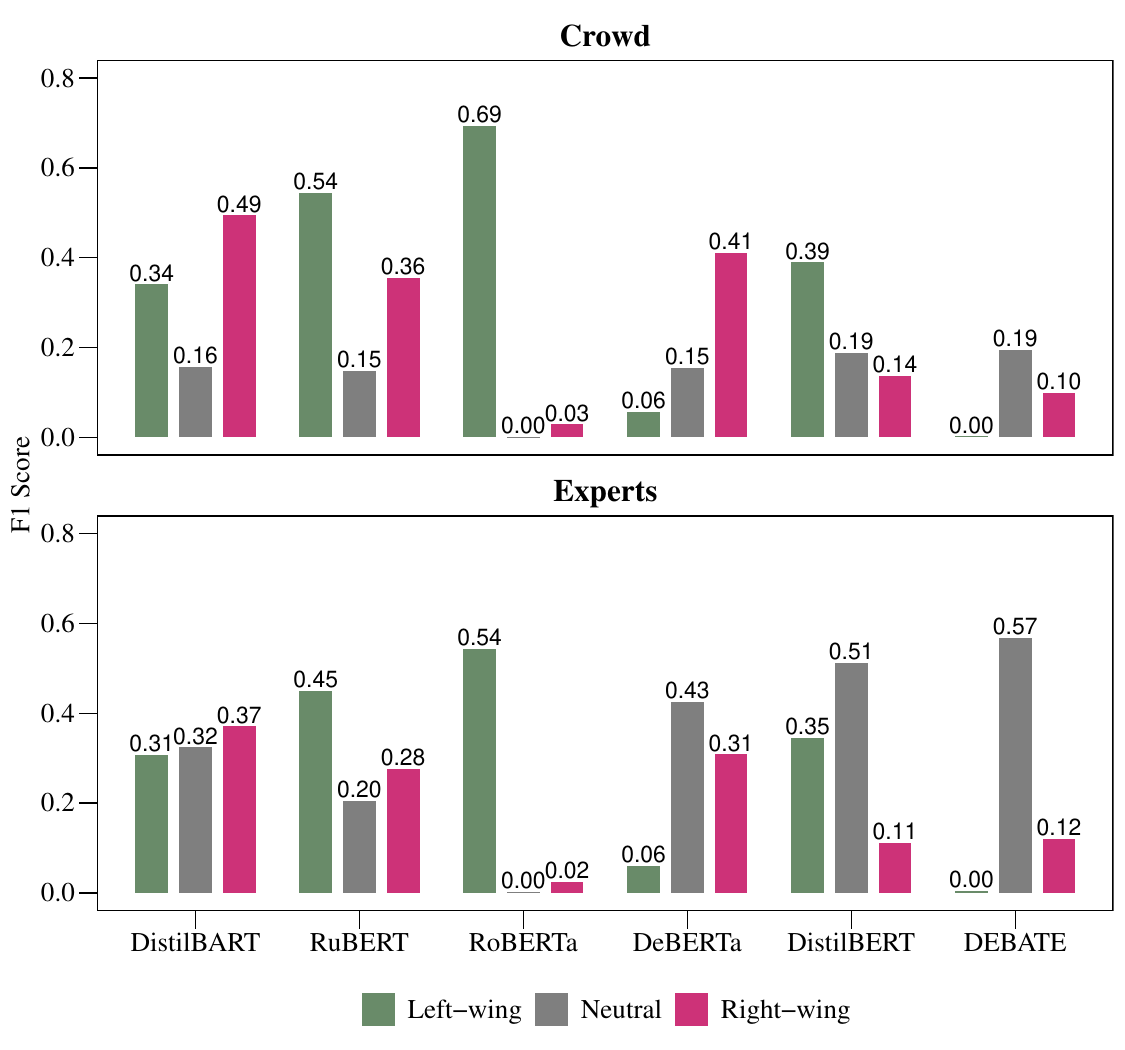}
\end{figure}

Compared to fine-tuning, zero-shot models offer researchers the opportunity to conduct automated text analyses without requiring pre-annotated data. This type of language models relies on its existing knowledge and language processing capabilities to execute new tasks that they were not trained on. I select models from the HuggingFace platform that were either top rated for zero-shot classification or built specifically for political text analysis. The classification pipeline starts with a hypothesis template that prompts the model about the economic ideology of each sentence. The model then selects one of the three candidate labels (left-wing, neutral, or right-wing). Similar to the previous analyses, I start by reporting sentence-level metrics before proceeding to overall correlations.

Figure \ref{2fig:zeroshot_f1_compare} and Table \ref{tab2:compare_0shot} show varying granular performance levels across zero-shot models. The DEBATE model, although specifically created for political documents processing, performs poorly against both crowd and expert benchmarks. Its F1 scores in the left-wing category are virtually zero. In the right-wing category, it returns scores of 0.12 and 0.1 against expert and crowd coders, respectively. In contrast, it shows a notable increase with neutral content when compared to experts (0.57). While RoBERTa achieves relatively fair F1 scores for left-leaning content (0.54 and 0.69). Its virtually zero scores for neutral and left-leaning content diminishes its viability as a reliable method of operationalizing economic ideology.

Compared to the other zero-shot models, DistilBERT and DeBERTa achieve average performance. While DeBERTa returns (relatively) fair scores for right-wing (0.31 for experts and 0.41 for crowd coders), it clearly performs poorly with left-wing content (0.06 against both references). DistilBERT shows opposite dynamics with good performance on left economic content (0.35 and 0.39) and poor metrics on right-wing sentences (0.11 and 0.14). Both models seem to match experts' assessment of neutrality but fail to mirror crowd coders in this category.

DistilBART and RuBERT achieve the highest scores in this category. DistilBART's reflection of expert judgments seems to be balanced \textemdash although fairly low \textemdash with F1 scores of 0.31, 0.32, and 0.37 for left, neutral, and right content. While these metrics improve when compared to crowd labeling for left and right content (0.34 and 0.49), its ability to capture neutral content declines. RuBERT shows similar metrics but with improved performance on left-wing content (0.45 and 0.54).

\begin{figure}[htb!]
	\centering
	\caption{Aggregate manifesto-level correlation coefficients of zero-shot models - expert coding.}
	\label{2fig:zeroshot_experts_corr}
	\includegraphics[width=0.8\linewidth]{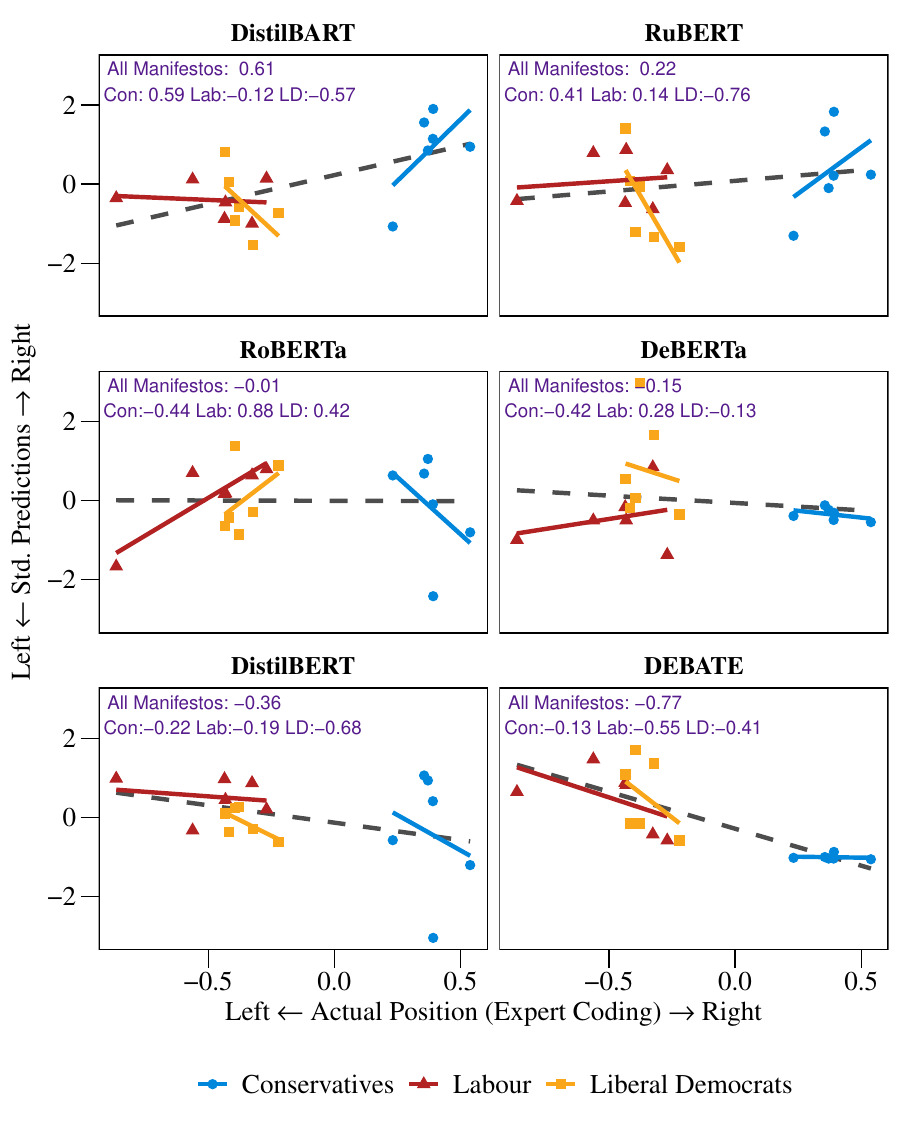}
\end{figure}

Evaluations of overall manifesto correlations show varying performance levels of zero-shot models. Using experts as a reference (Figure \ref{2fig:zeroshot_experts_corr}), DistilBART achieves the highest coefficient (0.61) which demonstrates relatively good capabilities in capturing the broad patterns of economic ideology. RuBERT follows with an overall correlation estimate of 0.22. The other models return negative correlation scores (RoBERTa -0.01; DeBERTa: -0.15; DistilBERT: -0.36). DEBATE's predictive abilities of ideology are particularly poor with a negative correlation estimate of -0.77. Not only does it fail to capture ideological labels, it also results in mismatched predictions.

For party-specific accuracy, DistilBART and RuBERT stand out in their ability to align with expert assessments of Conservative manifestos (0.59 and 0.41). However, they seem to struggle with Labour (-0.12 and 0.14) and Liberal Democrats manifestos (-0.57 and -0.76). Other models maintain the negative values observed in the overall correlation scores. A notable observation, however, is RoBERTa's improved performance with Labour (0.88) and Liberal Democrats manifestos (0.42).

Compared against crowd coders' assessments (Figure \ref{2fig:zeroshot_crowd_corr}), the correlation coefficients for DistilBART and RuBERT slightly improve. In contrast, the other models either maintain the negative relationship or return even poorer estimates. Within-party correlations show similar dynamics as the ones observed against expert labeling. These results underscore the challenges that zero-shot models face when dealing with tasks without prior training. While they offer flexibility and time-savings, they lack in accuracy and reliability. Ideological scaling is a complex task that requires understanding domain- and context-specific terminologies and semantics. Relying solely on pre-existing knowledge to measure economic ideology comes with several caveats.

\subsection*{Prompting Zero-shot Models}
Using DistilBART, the top performing zero-shot model, I assess how different prompting strategies affect model performance. I formulate four prompts variations with different levels of details. Prompts 1 and 2 provide clear definitions of what each ideological label entails. Prompts 3 and 4 simply query the model about the sentence's category without any further explanations. Prompts 2 and 4 include an added layer of instruction asking the model to take into account both implicit and explicit signals of economic ideology. These prompt variations are as follows:

\begin{singlespace}
	\begin{quotation}
		Prompt 1: ``Right-wing beliefs emphasize free-market capitalism, low taxes, free trade, deregulation, privatization, individualism, promoting the private sector, and limited government intervention. Left-wing beliefs emphasize government intervention, wealth redistribution, protectionism, progressive taxation, expanded welfare programs, and government regulation. Neutral refers to apolitical or factual content. The political economic ideology expressed in this statement is ...''
	\end{quotation}
	
	\begin{quotation}
		Prompt 2: ``Right-wing beliefs emphasize free-market capitalism, low taxes, free trade, deregulation, privatization, individualism, promoting the private sector, and limited government intervention. Left-wing beliefs emphasize government intervention, wealth redistribution, protectionism, progressive taxation, expanded welfare programs, and government regulation. Neutral refers to apolitical or factual content. The political economic ideology expressed \textbf{(explicitly or implicitly)} in this statement is ..''
	\end{quotation}
	
	\begin{quotation}
		Prompt 3: ``The political economic ideology expressed in this statement is ...''
	\end{quotation}
	
	\begin{quotation}
		Prompt 4: ``The political economic ideology expressed \textbf{(explicitly or implicitly)} in this statement is ...''
	\end{quotation}	
\end{singlespace}

\begin{figure}[htb!]
	\centering
	\caption{Performance metrics of zero-shot DistilBART using different prompt variations.}
	\label{2fig:prompt_f1_compare}
	\includegraphics[width=0.8\linewidth]{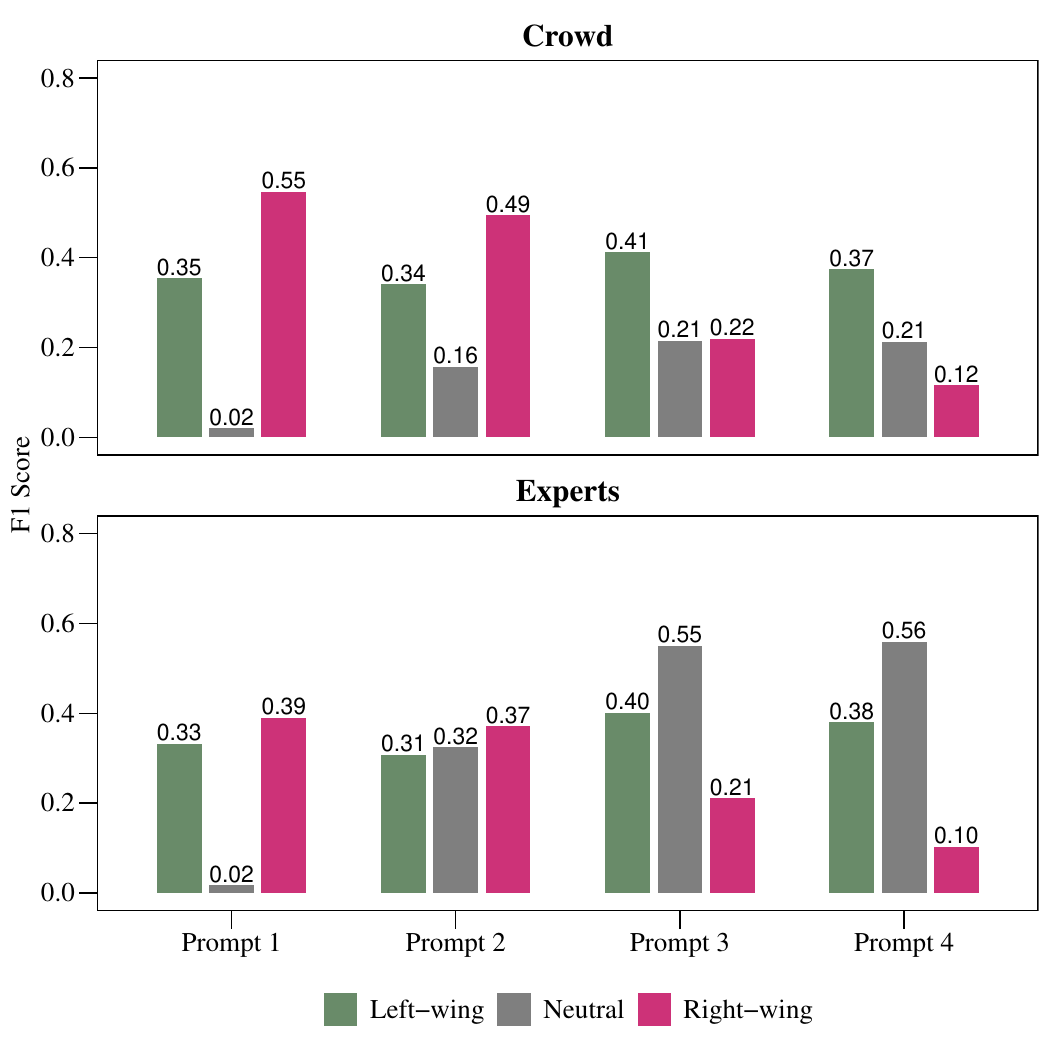}
\end{figure}

Figure \ref{2fig:prompt_f1_compare} and Table \ref{tab2:compare_prompts} summarize the performance of zero-shot DistilBART across four prompt variations. Starting with the expert benchmark, prompt 3 results in the highest F1 scores for the left-wing and neutral labels (0.4 and 0.55 against experts; 0.41 and 0.21 against crowd coders). However, it is outperformed by prompts 1 and 2 for right-wing content: F1 scores of 0.39 and 0.37 compared to experts; 0.55 and 0.49 compared to crowd labeling. These trends suggest that while the model is better at detecting left and neutral content without explicit definitions, it benefits greatly from prompts explaining right-wing economic ideology and its beliefs. Prompt 4 follows similar dynamics to prompt 3 but with lower performance indicators. Moving to the crowd assessments benchmark, Figure \ref{2fig:prompt_f1_compare} shows varying changes in F1 scores. Notable increases include the right-wing category for prompt 1 (from 0.39 to 0.55). In contrast, the metrics for the neutral categories decline across all models. Another key takeaway is the impact of asking the model to look for implicit cues. Prompts that included this instruction yielded lower F1 scores. This is likely due to the added ambiguity that comes with latent ideological signals.
 
\begin{figure}[htb!]
	\centering
	\caption{Aggregate Manifesto-level correlation coefficients of zero-shot DistilBART using different prompt variations - expert coding.}
	\label{2fig:prompt_experts_corr}
	\includegraphics[width=0.8\linewidth]{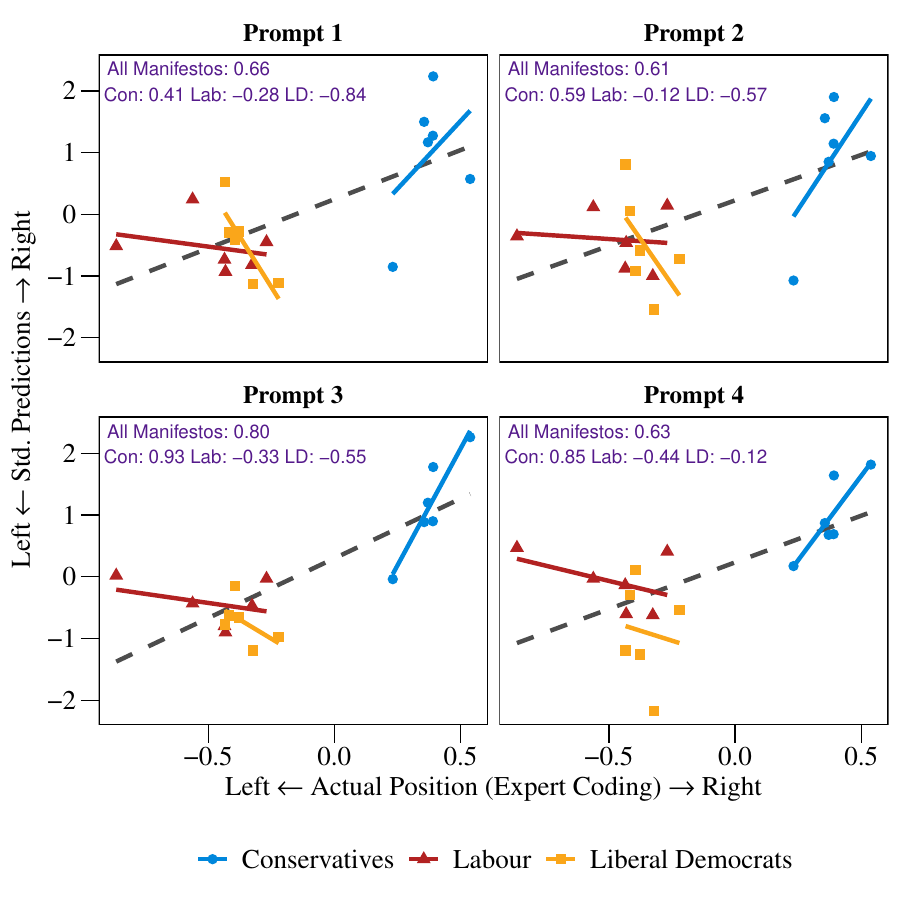}
\end{figure}

Similar to previous analyses, I also investigate how different prompts perform at the manifesto level. The results in Figure \ref{2fig:prompt_experts_corr} follow the same dynamics as the sentence-level evaluation. Prompt 3, which did not include explicit definitions or instructions to look for indirect signals, achieved the highest overall coefficient (0.8). In contrast, the F1 scores for prompt 1 is 0.66. This suggests that providing explanations and definitions might limit the model's scope. Brief and concise prompts that explicitly query the model result in better detection of overarching ideological trends. These estimates also confirm the takeaway from the sentence-level evaluation. Prompts that did not direct the model to account for subtle ideological references achieved higher coefficients. The coefficients of correlation with crowd coding show similar dynamics with varying degrees in estimate improvement (Figure \ref{2fig:prompt_crowd_corr}).

The implications from this prompt comparison are twofold. First, zero-shot models appear to react differently to prompt variations depending on the ideological label. For some labels, concise and explicit prompts achieve better results. For others, the models might benefit from provided guidelines. Second, instructing zero-shot models to take implicit expressions of economic ideology into account decreases performance. This is potentially due to the added layer of complexity inherent in indirect ideological references. Overall, prompt formulation appears to be class and task dependent.

\section{Discussion}
Generative language models consistently outperform fine-tuned and zero-shot approaches to measuring economic ideology in manifesto sentences. Particularly, GPT-4o provides the closest alignment with both experts and crowd coders. It achieves the best performance metrics at sentence and manifesto levels. However, it is only accessible through the paid OpenAI developer platform. Gemini 1.5 Flash achieves slightly lower but comparable results across all benchmarks. It also has the benefit of being available for free, as of the time of writing this chapter. However, it is not guaranteed that Google will maintain it as a free service. This underscores the importance of open-source alternatives, that also have the potential to become more computationally efficient, as discussed by \citet{spirling_2023}.

Fine-tuned language models offer an alternative option to generative LLMs. The results show that they achieve competitive results and are often freely accessible. Not requiring advanced (and expensive) hardware such as GPUs also contributes to this resource sensitivity. Their inference time is considerably faster which is not only beneficial for research purposes but also serves ethical aims as they are computationally efficient. However, they may be limited in scalability as they require pre-annotated data and face challenges with transfer learning. These advantages and drawbacks further highlight the trade-offs involved with model selection.

Zero-shot models fail to accurately capture trends and labels of economic ideology in manifesto sentences. They yield considerably lower performance compared to generative and fine-tuning approaches. The absence of task-specific optimization renders their output inaccurate, measured both at the sentence and manifesto levels. Prompt experimentation may improve their performance. The analysis indicates that class-based metrics require different prompting styles. For some ideological labels, concise prompts that only provide the specific instruction lead to better results. For other categories, the models may benefit from clear guidelines and definitions. Nonetheless, the results from the best performing prompt fail to match generative and fine-tuned models. Zero-shot approaches may offer flexibility and convenience but come with several caveats in terms of efficacy.

The analyses also highlight that researchers should rely on both sentence-level and aggregate approaches to validate measurements of economic ideology. While F1 scores offer detailed references for performance, manifesto-level correlations provide a broader picture of alignment with ideological dynamics. For several applications in political science research, overall correlations may be a more suitable measure of economic ideology. While some applications require document- or sentence-level accuracy, most studies would benefit more from better overarching alignment. However, researchers should account for the fact that holistic approaches average out granular inconsistencies. \citet{mullerproksh2024} discuss that, regardless of methods' performance, aggregation results in higher similarities with human coding. Granular sentence classifications may show individual variability but aggregation usually paints a broader picture that diminishes these inconsistencies.

Another important finding is the varying predictive metrics when compared to the two human-coded benchmarks. The analyses show that most models' classification of economic ideology better aligns with experts than crowd coders. The \citet{benoit_crowd-sourced_2016} replication data employed academic researchers, including the authors, as expert annotators. This suggests that their ratings of manifestos are arguably more standardized and objective. In contrast, crowd coders may be more subjective and likely to insert biases into their annotations. These challenges emphasize the need to adopt robust guidelines for the annotation of political content.

Manifesto-level correlation metrics reveal considerable within-party differences in performance. For instance, generative models such as GPT-4o reliably reflect expert judgments for Conservative and Labour manifestos. However, they struggle with the more moderate positions of Liberal Democrats. This suggests that models' capabilities decline when faced with less ideological economic content. Considering these variations goes through careful experimentation as well as country- and party- specific validation.
\section{Conclusion}
This study offers one the few systematic comparisons of language models' capabilities in operationalizing (economic) ideology. Focusing on sentences on economic policy from U.K. party manifestos spanning six elections, I rely on expert and crowd annotations as evaluation benchmarks of models' predictions. The analyses compare three major categories of language models: generative, fine-tuned, and zero-shot. By calculating sentence-level and aggregated validation measures, the results provide insights into each model's capabilities and limitations.

Generative models proved to be the most effective in detecting economic ideology across all benchmarks and validation approaches. GPT-4o and Gemini 1.5 Flash show excellent predictive and adaptive capabilities. These models, however, require considerable financial and computational resources and they usually rely on proprietary access. Alternatives include open-source models but also fine-tuning approaches.

Fine-tuned models offered a balance between performance and accessibility. Although lower than generative models, fine-tuning yielded promising and competitive metrics. The results also highlight their dependency on training data as they face difficulties with transfer learning. Political science research often requires the examination of documents across different institutional, social, and cultural contexts. Requiring unique training sets for different data types limits scalability. Experimenting with different sizes of the training data showed that fine-tuning is more efficient at capturing economic ideology with larger sets.

Zero-shot models fell short of expectations as they repeatedly struggled with identifying economic ideology labels. Applying general-purpose approaches to the domain-specific task of ideology classification proved to be unreliable. Prompt experimentation showed that the performance of zero-shot models using different prompts could improve but remains heavily dependent on and variable across the target categories.

Other key findings include within-party variation of models' predictive capabilities. Several models performed poorly on manifestos of the centrist Liberal Democrats, potentially due to the absence of clear ideological cues in moderate content. In addition, the analysis shows that most models align more with expert labeling than crowd annotations. This is likely due to their inherent objective and systematic approaches to executing tasks.

In summary, this paper contributes to the computational literature with a systematic review of language models' effectiveness at detecting economic ideology, a key concept in political science. In addition to providing guidelines and best-practices, it lays a foundation for future research relying on automated scaling of ideology. The assessments carried out in the analysis point to each model's strengths and limitations. With careful reviews of these trade-offs, researchers can select models that best fit their project-specific objectives.

\pagebreak
\singlespacing
\bibliographystyle{apalike}

\bibliography{refs.bib}
	\pagebreak
\appendix
\begin{center}
	\LARGE
	Supporting Information
\end{center}

\begin{figure}[htbp!]
	\centering
	\caption{Aggregate manifesto-level correlation coefficients by model - crowd coding.}
	\label{2fig:overview_crowd}
	\includegraphics[width=1\linewidth]{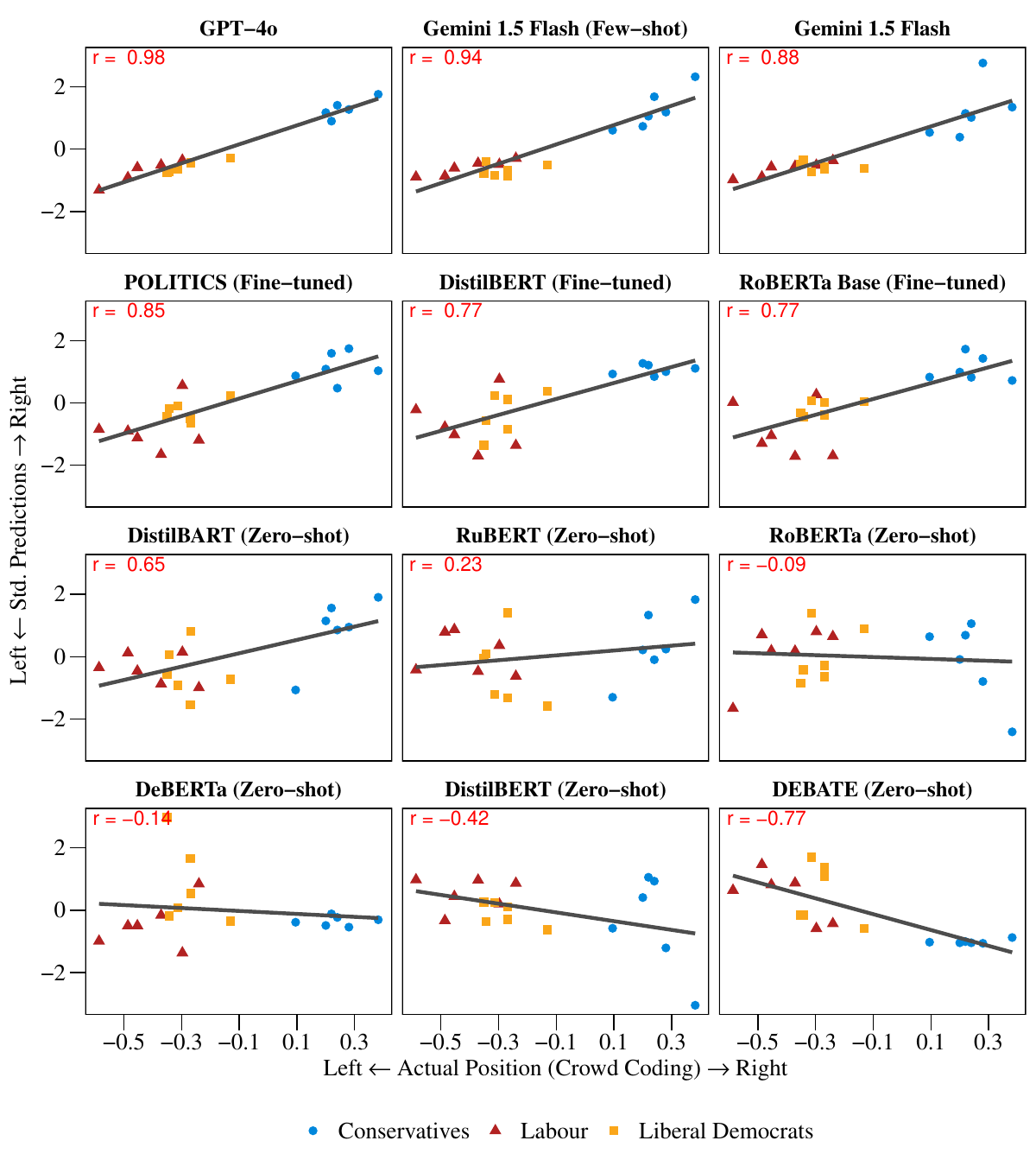}
\end{figure}

% latex table generated in R 4.3.1 by xtable 1.8-4 package
% Wed Dec  4 18:24:08 2024
\begin{table}[ht!]
\centering
\caption{Performance metrics of generative models in detecting economic ideology.} 
\label{tab2:compare_gen}
\begingroup\footnotesize
\begin{tabular}{llrrrrl}
  \hline
Classifier & Class & F1 & Accuracy & Precision & Recall & Source \\ 
  \hline
GPT-4o & left-wing & 0.56 & 0.64 & 0.85 & 0.42 & Crowd \\ 
  GPT-4o & neutral or procedural & 0.28 & 0.64 & 0.17 & 0.84 & Crowd \\ 
  GPT-4o & right-wing & 0.56 & 0.67 & 0.83 & 0.42 & Crowd \\ 
  Gemini 1.5 Flash & left-wing & 0.71 & 0.64 & 0.71 & 0.71 & Crowd \\ 
  Gemini 1.5 Flash & neutral or procedural & 0.24 & 0.56 & 0.18 & 0.37 & Crowd \\ 
  Gemini 1.5 Flash & right-wing & 0.55 & 0.65 & 0.69 & 0.46 & Crowd \\ 
  Gemini 1.5 Flash 
(Few-shot) & left-wing & 0.71 & 0.64 & 0.72 & 0.70 & Crowd \\ 
  Gemini 1.5 Flash 
(Few-shot) & neutral or procedural & 0.28 & 0.60 & 0.19 & 0.49 & Crowd \\ 
  Gemini 1.5 Flash 
(Few-shot) & right-wing & 0.53 & 0.65 & 0.71 & 0.43 & Crowd \\ 
  GPT-4o & left-wing & 0.66 & 0.72 & 0.75 & 0.59 & Experts \\ 
  GPT-4o & neutral or procedural & 0.64 & 0.68 & 0.56 & 0.74 & Experts \\ 
  GPT-4o & right-wing & 0.68 & 0.77 & 0.75 & 0.62 & Experts \\ 
  Gemini 1.5 Flash & left-wing & 0.68 & 0.68 & 0.57 & 0.83 & Experts \\ 
  Gemini 1.5 Flash & neutral or procedural & 0.41 & 0.58 & 0.60 & 0.31 & Experts \\ 
  Gemini 1.5 Flash & right-wing & 0.63 & 0.74 & 0.61 & 0.66 & Experts \\ 
  Gemini 1.5 Flash 
(Few-shot) & left-wing & 0.68 & 0.69 & 0.58 & 0.82 & Experts \\ 
  Gemini 1.5 Flash 
(Few-shot) & neutral or procedural & 0.46 & 0.60 & 0.61 & 0.37 & Experts \\ 
  Gemini 1.5 Flash 
(Few-shot) & right-wing & 0.64 & 0.74 & 0.64 & 0.63 & Experts \\ 
   \hline
\end{tabular}
\endgroup
\end{table}

\begin{figure}[ht!]
	\centering
	\caption{Aggregate manifesto-level correlation coefficients of generative models - crowd coding.}
	\label{2fig:gen_crowd_corr}
	\includegraphics[width=1\linewidth]{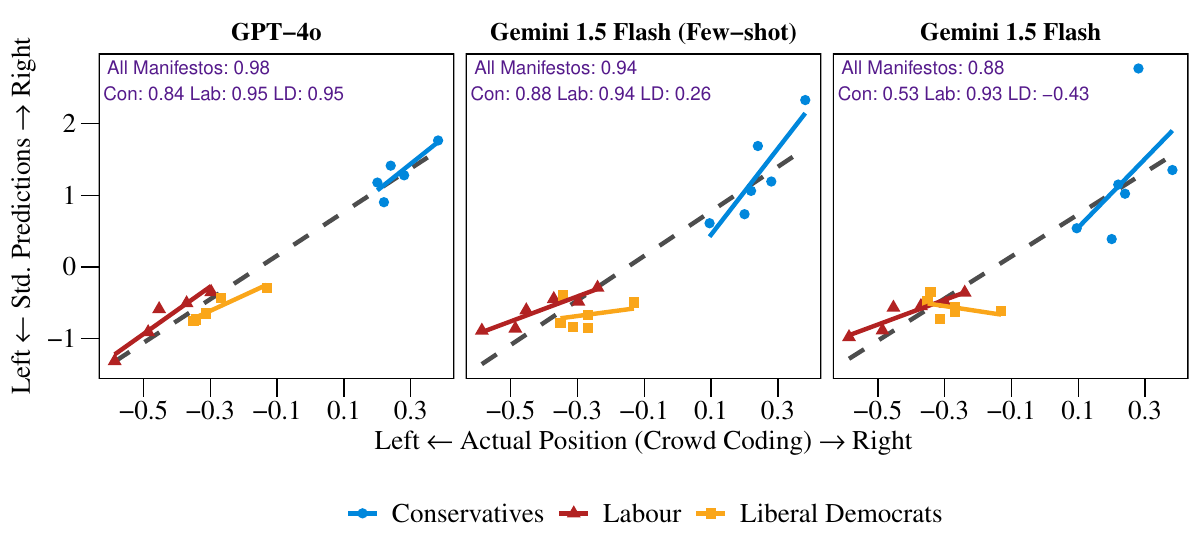}
\end{figure}

% latex table generated in R 4.3.1 by xtable 1.8-4 package
% Wed Dec  4 18:36:13 2024
\begin{table}[ht!]
\centering
\caption{Manifesto-level predicted economic ideology scores by generative models.} 
\label{tab2:manif_crowd}
\begingroup\footnotesize
\begin{tabular}{lrrrrr}
  \hline
Manifesto & Crowd Pos. & Experts Pos. & GPT-4o & Gemini 1.5 Flash & Gemini 1.5 Flash (few-shot) \\ 
  \hline
Con 1987 & 0.20 & 0.39 & 1.32 & 1.28 & 1.26 \\ 
  Con 1992 & 0.22 & 0.36 & 1.23 & 1.96 & 1.45 \\ 
  Con 1997 & 0.24 & 0.37 & 1.40 & 1.84 & 1.83 \\ 
  Con 2001 & 0.28 & 0.54 & 1.35 & 3.39 & 1.53 \\ 
  Con 2005 & 0.38 & 0.39 & 1.51 & 2.14 & 2.21 \\ 
  Con 2010 & 0.10 & 0.23 & 1.00 & 1.42 & 1.18 \\ 
  Lab 1987 & -0.59 & -0.86 & 0.50 & 0.07 & 0.29 \\ 
  Lab 1992 & -0.49 & -0.56 & 0.63 & 0.15 & 0.31 \\ 
  Lab 1997 & -0.30 & -0.27 & 0.82 & 0.48 & 0.53 \\ 
  Lab 2001 & -0.45 & -0.43 & 0.74 & 0.43 & 0.46 \\ 
  Lab 2005 & -0.37 & -0.44 & 0.77 & 0.45 & 0.55 \\ 
  Lab 2010 & -0.24 & -0.33 & 1.00 & 0.62 & 0.65 \\ 
  LD 1987 & -0.31 & -0.39 & 0.72 & 0.29 & 0.32 \\ 
  LD 1992 & -0.35 & -0.38 & 0.68 & 0.51 & 0.35 \\ 
  LD 1997 & -0.27 & -0.43 & 0.79 & 0.38 & 0.31 \\ 
  LD 2001 & -0.34 & -0.42 & 0.69 & 0.63 & 0.58 \\ 
  LD 2005 & -0.13 & -0.22 & 0.84 & 0.39 & 0.52 \\ 
  LD 2010 & -0.27 & -0.32 & 1.00 & 0.43 & 0.42 \\ 
   \hline
\end{tabular}
\endgroup
\end{table}

% latex table generated in R 4.3.1 by xtable 1.8-4 package
% Tue Dec  3 18:57:21 2024
\begin{table}[ht!]
\centering
\caption{Performance metrics of fine-tuned models in detecting economic ideology.} 
\label{tab2:compare_finetuned}
\begingroup\footnotesize
\begin{tabular}{lllrrrr}
  \hline
Classifier & Source & Class & F1 & Accuracy & Precision & Recall \\ 
  \hline
POLITICS & Crowd & left-wing & 0.67 & 0.59 & 0.73 & 0.62 \\ 
  POLITICS & Crowd & neutral or procedural & 0.25 & 0.58 & 0.16 & 0.59 \\ 
  POLITICS & Crowd & right-wing & 0.39 & 0.59 & 0.69 & 0.27 \\ 
  DistilBERT & Crowd & left-wing & 0.69 & 0.59 & 0.67 & 0.71 \\ 
  DistilBERT & Crowd & neutral or procedural & 0.24 & 0.55 & 0.19 & 0.34 \\ 
  DistilBERT & Crowd & right-wing & 0.45 & 0.59 & 0.57 & 0.37 \\ 
  RoBERTa Base & Crowd & left-wing & 0.66 & 0.54 & 0.72 & 0.61 \\ 
  RoBERTa Base & Crowd & neutral or procedural & 0.23 & 0.56 & 0.14 & 0.64 \\ 
  RoBERTa Base & Crowd & right-wing & 0.19 & 0.54 & 0.73 & 0.11 \\ 
  POLITICS & Experts & left-wing & 0.64 & 0.68 & 0.55 & 0.77 \\ 
  POLITICS & Experts & neutral or procedural & 0.56 & 0.61 & 0.61 & 0.52 \\ 
  POLITICS & Experts & right-wing & 0.49 & 0.65 & 0.60 & 0.42 \\ 
  DistilBERT & Experts & left-wing & 0.60 & 0.61 & 0.47 & 0.82 \\ 
  DistilBERT & Experts & neutral or procedural & 0.38 & 0.56 & 0.64 & 0.27 \\ 
  DistilBERT & Experts & right-wing & 0.45 & 0.61 & 0.44 & 0.46 \\ 
  RoBERTa Base & Experts & left-wing & 0.64 & 0.68 & 0.55 & 0.77 \\ 
  RoBERTa Base & Experts & neutral or procedural & 0.58 & 0.60 & 0.56 & 0.60 \\ 
  RoBERTa Base & Experts & right-wing & 0.29 & 0.57 & 0.71 & 0.18 \\ 
   \hline
\end{tabular}
\endgroup
\end{table}

\begin{figure}[ht!]
	\centering
	\caption{Aggregate manifesto-level correlation coefficients of fine-tuned models - crowd coding.}
	\label{2fig:finetune_crowd_corr}
	\includegraphics[width=1\linewidth]{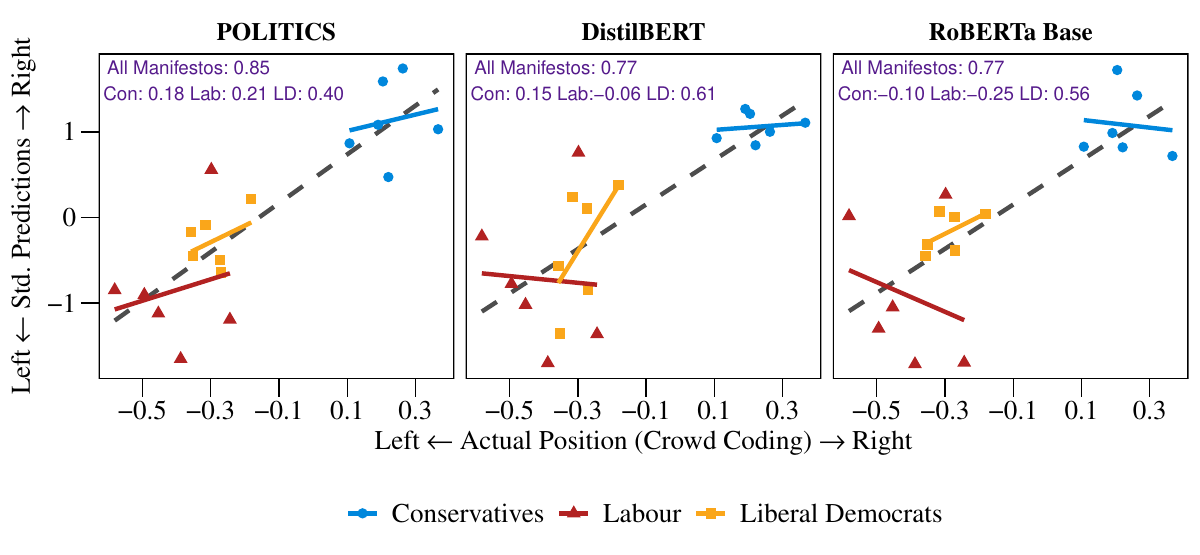}
\end{figure}

% latex table generated in R 4.3.1 by xtable 1.8-4 package
% Tue Dec  3 19:03:31 2024
\begin{table}[ht!]
\centering
\caption{Manifesto-level predicted economic ideology scores by fine-tuned models.} 
\label{tab2:manif_crowd_finetuned}
\begingroup\footnotesize
\begin{tabular}{lrrrrr}
  \hline
Manifesto & Crowd Pos. & Experts Pos. & POLITICS & RoBERTa & DistilBERT \\ 
  \hline
Con 1987 & 0.19 & 0.37 & 0.87 & 0.70 & 0.93 \\ 
  Con 1992 & 0.20 & 0.33 & 0.91 & 0.77 & 0.92 \\ 
  Con 1997 & 0.22 & 0.33 & 0.82 & 0.68 & 0.90 \\ 
  Con 2001 & 0.26 & 0.51 & 0.93 & 0.74 & 0.91 \\ 
  Con 2005 & 0.37 & 0.36 & 0.87 & 0.67 & 0.91 \\ 
  Con 2010 & 0.11 & 0.22 & 0.85 & 0.68 & 0.90 \\ 
  Lab 1987 & -0.58 & -0.87 & 0.70 & 0.60 & 0.82 \\ 
  Lab 1992 & -0.49 & -0.57 & 0.70 & 0.46 & 0.78 \\ 
  Lab 1997 & -0.30 & -0.26 & 0.82 & 0.62 & 0.89 \\ 
  Lab 2001 & -0.45 & -0.43 & 0.68 & 0.49 & 0.77 \\ 
  Lab 2005 & -0.39 & -0.43 & 0.63 & 0.42 & 0.72 \\ 
  Lab 2010 & -0.24 & -0.33 & 0.67 & 0.42 & 0.74 \\ 
  LD 1987 & -0.32 & -0.40 & 0.77 & 0.60 & 0.85 \\ 
  LD 1992 & -0.35 & -0.38 & 0.74 & 0.56 & 0.74 \\ 
  LD 1997 & -0.27 & -0.44 & 0.72 & 0.56 & 0.78 \\ 
  LD 2001 & -0.36 & -0.43 & 0.76 & 0.55 & 0.80 \\ 
  LD 2005 & -0.18 & -0.26 & 0.79 & 0.60 & 0.86 \\ 
  LD 2010 & -0.27 & -0.31 & 0.73 & 0.60 & 0.84 \\ 
   \hline
\end{tabular}
\endgroup
\end{table}

% latex table generated in R 4.3.1 by xtable 1.8-4 package
% Tue Dec  3 19:54:24 2024
\begin{table}[ht!]
\centering
\caption{Performance metrics of fine-tuned POLITICS model at different sizes of the training set.} 
\label{tab2:finetuned_optimize}
\begingroup\footnotesize
\begin{tabular}{lllrrrr}
  \hline
N-Sentences & Source & Class & F1 & Accuracy & Precision & Recall \\ 
  \hline
1000 & Crowd & left-wing & 0.67 & 0.59 & 0.73 & 0.62 \\ 
  1000 & Crowd & neutral or procedural & 0.25 & 0.58 & 0.16 & 0.59 \\ 
  1000 & Crowd & right-wing & 0.39 & 0.59 & 0.69 & 0.27 \\ 
  900 & Crowd & left-wing & 0.60 & 0.64 & 0.78 & 0.49 \\ 
  900 & Crowd & neutral or procedural & 0.24 & 0.54 & 0.17 & 0.40 \\ 
  900 & Crowd & right-wing & 0.57 & 0.62 & 0.54 & 0.61 \\ 
  800 & Crowd & left-wing & 0.64 & 0.64 & 0.75 & 0.56 \\ 
  800 & Crowd & neutral or procedural & 0.27 & 0.57 & 0.22 & 0.34 \\ 
  800 & Crowd & right-wing & 0.58 & 0.63 & 0.53 & 0.64 \\ 
  700 & Crowd & left-wing & 0.61 & 0.62 & 0.75 & 0.52 \\ 
  700 & Crowd & neutral or procedural & 0.24 & 0.54 & 0.18 & 0.35 \\ 
  700 & Crowd & right-wing & 0.54 & 0.60 & 0.50 & 0.59 \\ 
  600 & Crowd & left-wing & 0.69 & 0.54 & 0.64 & 0.76 \\ 
  600 & Crowd & neutral or procedural & 0.23 & 0.53 & 0.16 & 0.36 \\ 
  600 & Crowd & right-wing & 0.30 & 0.55 & 0.59 & 0.20 \\ 
  1000 & Experts & left-wing & 0.66 & 0.68 & 0.59 & 0.76 \\ 
  1000 & Experts & neutral or procedural & 0.57 & 0.63 & 0.59 & 0.54 \\ 
  1000 & Experts & right-wing & 0.48 & 0.65 & 0.61 & 0.40 \\ 
  900 & Experts & left-wing & 0.65 & 0.68 & 0.66 & 0.64 \\ 
  900 & Experts & neutral or procedural & 0.48 & 0.60 & 0.65 & 0.38 \\ 
  900 & Experts & right-wing & 0.52 & 0.68 & 0.41 & 0.73 \\ 
  800 & Experts & left-wing & 0.65 & 0.66 & 0.61 & 0.69 \\ 
  800 & Experts & neutral or procedural & 0.41 & 0.59 & 0.71 & 0.29 \\ 
  800 & Experts & right-wing & 0.51 & 0.66 & 0.39 & 0.72 \\ 
  700 & Experts & left-wing & 0.60 & 0.62 & 0.59 & 0.61 \\ 
  700 & Experts & neutral or procedural & 0.41 & 0.57 & 0.62 & 0.31 \\ 
  700 & Experts & right-wing & 0.48 & 0.64 & 0.37 & 0.69 \\ 
  600 & Experts & left-wing & 0.60 & 0.59 & 0.48 & 0.83 \\ 
  600 & Experts & neutral or procedural & 0.42 & 0.55 & 0.56 & 0.33 \\ 
  600 & Experts & right-wing & 0.30 & 0.55 & 0.46 & 0.22 \\ 
   \hline
\end{tabular}
\endgroup
\end{table}

\begin{figure}[ht!]
	\centering
	\caption{Aggregate manifesto-level correlation coefficients of fine-tuned POLITICS at different sizes of the training set - crowd coding.}
	\label{2fig:optimize_crowd_corr}
	\includegraphics[width=0.8\linewidth]{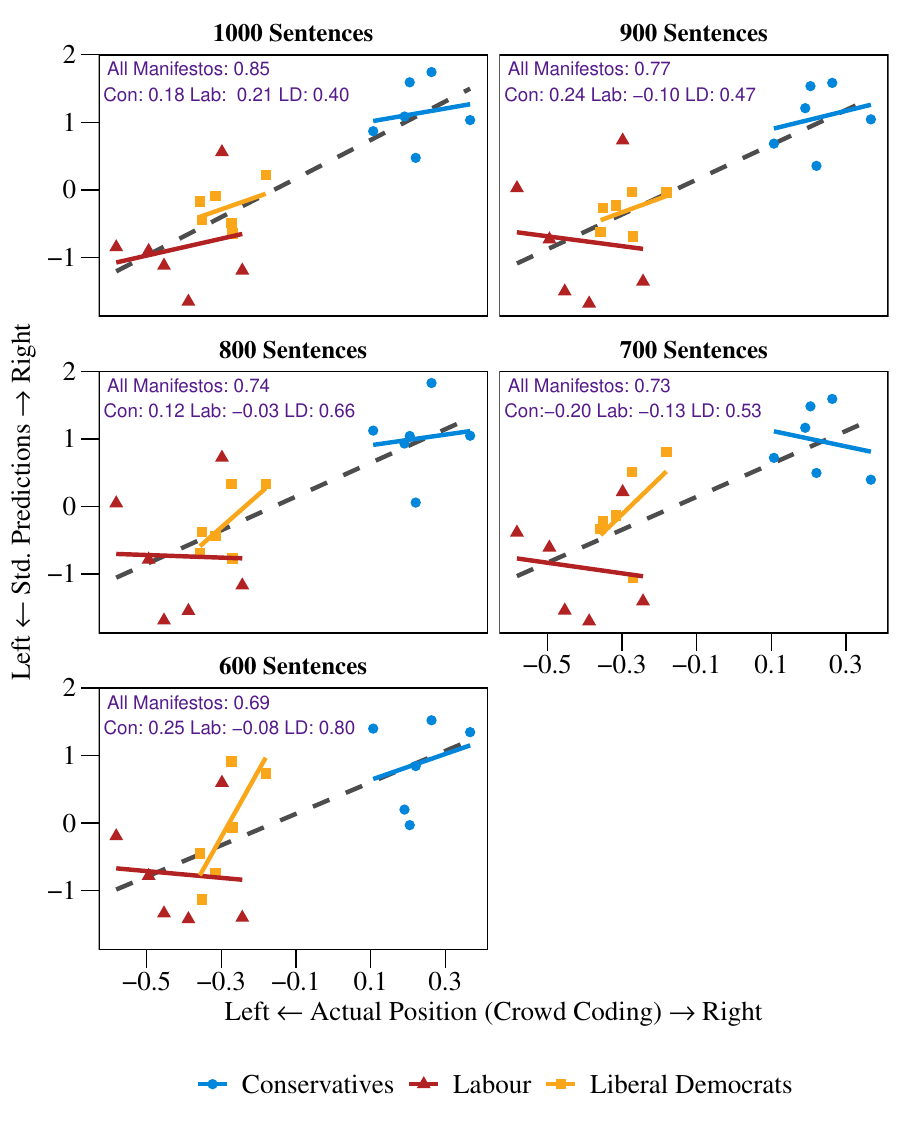}
\end{figure}
% latex table generated in R 4.3.1 by xtable 1.8-4 package
% Wed Nov  6 14:48:20 2024
\begin{table}[ht!]
\centering
\caption{Manifesto-level predicted economic ideology scores using different prompts (DistilBART model).} 
\label{tab2:prompt_crowd}
\begingroup\footnotesize
\begin{tabular}{lrrrrrr}
  \hline
Manifesto & Crowd Pos. & Experts Pos. & Prompt 1 & Prompt 2 & Prompt 3 & Prompt 4 \\ 
  \hline
Con 1987 & 0.20 & 0.39 & 1.49 & 1.38 & 0.94 & 0.80 \\ 
  Con 1992 & 0.22 & 0.36 & 1.51 & 1.42 & 0.94 & 0.83 \\ 
  Con 1997 & 0.24 & 0.37 & 1.48 & 1.35 & 0.97 & 0.80 \\ 
  Con 2001 & 0.28 & 0.54 & 1.41 & 1.36 & 1.08 & 0.96 \\ 
  Con 2005 & 0.38 & 0.39 & 1.60 & 1.46 & 1.03 & 0.94 \\ 
  Con 2010 & 0.10 & 0.23 & 1.25 & 1.15 & 0.85 & 0.72 \\ 
  Lab 1987 & -0.59 & -0.86 & 1.29 & 1.22 & 0.86 & 0.77 \\ 
  Lab 1992 & -0.49 & -0.56 & 1.37 & 1.27 & 0.81 & 0.69 \\ 
  Lab 1997 & -0.30 & -0.27 & 1.30 & 1.27 & 0.85 & 0.76 \\ 
  Lab 2001 & -0.45 & -0.43 & 1.24 & 1.21 & 0.76 & 0.61 \\ 
  Lab 2005 & -0.37 & -0.44 & 1.26 & 1.17 & 0.77 & 0.68 \\ 
  Lab 2010 & -0.24 & -0.33 & 1.26 & 1.15 & 0.81 & 0.61 \\ 
  LD 1987 & -0.31 & -0.39 & 1.30 & 1.16 & 0.84 & 0.71 \\ 
  LD 1992 & -0.35 & -0.38 & 1.32 & 1.20 & 0.79 & 0.51 \\ 
  LD 1997 & -0.27 & -0.43 & 1.40 & 1.34 & 0.78 & 0.52 \\ 
  LD 2001 & -0.34 & -0.42 & 1.31 & 1.26 & 0.79 & 0.65 \\ 
  LD 2005 & -0.13 & -0.22 & 1.22 & 1.18 & 0.76 & 0.62 \\ 
  LD 2010 & -0.27 & -0.32 & 1.22 & 1.10 & 0.73 & 0.38 \\ 
   \hline
\end{tabular}
\endgroup
\end{table}

% latex table generated in R 4.3.1 by xtable 1.8-4 package
% Wed Nov  6 14:44:32 2024
\begin{table}[ht!]
\centering
\caption{Performance metrics of zero-shot models in detecting economic ideology.} 
\label{tab2:compare_0shot}
\begingroup\footnotesize
\begin{tabular}{lllrrrr}
  \hline
Classifier & Source & Class & F1 & Accuracy & Prevision & Recall \\ 
  \hline
DistilBART & Crowd & Left-wing & 0.34 & 0.52 & 0.68 & 0.23 \\ 
  DistilBART & Crowd & Neutral & 0.16 & 0.44 & 0.11 & 0.27 \\ 
  DistilBART & Crowd & Right-wing & 0.49 & 0.47 & 0.41 & 0.63 \\ 
  DeBERTa & Crowd & Left-wing & 0.06 & 0.49 & 0.54 & 0.03 \\ 
  DeBERTa & Crowd & Neutral & 0.15 & 0.37 & 0.10 & 0.35 \\ 
  DeBERTa & Crowd & Right-wing & 0.41 & 0.33 & 0.33 & 0.54 \\ 
  DistilBERT & Crowd & Left-wing & 0.39 & 0.38 & 0.58 & 0.29 \\ 
  DistilBERT & Crowd & Neutral & 0.19 & 0.45 & 0.11 & 0.66 \\ 
  DistilBERT & Crowd & Right-wing & 0.14 & 0.43 & 0.32 & 0.09 \\ 
  RuBERT & Crowd & Left-wing & 0.54 & 0.46 & 0.55 & 0.54 \\ 
  RuBERT & Crowd & Neutral & 0.15 & 0.47 & 0.13 & 0.17 \\ 
  RuBERT & Crowd & Right-wing & 0.36 & 0.47 & 0.37 & 0.34 \\ 
  RoBERTa & Crowd & Left-wing & 0.69 & 0.50 & 0.53 & 0.99 \\ 
  RoBERTa & Crowd & Neutral & 0.00 & 0.50 & 0.08 & 0.00 \\ 
  RoBERTa & Crowd & Right-wing & 0.03 & 0.50 & 0.44 & 0.01 \\ 
  DEBATE & Crowd & Left-wing & 0.00 & 0.50 & 1.00 & 0.00 \\ 
  DEBATE & Crowd & Neutral & 0.19 & 0.51 & 0.11 & 0.99 \\ 
  DEBATE & Crowd & Right-wing & 0.10 & 0.50 & 0.78 & 0.05 \\ 
  DistilBART & Experts & Left-wing & 0.31 & 0.48 & 0.47 & 0.23 \\ 
  DistilBART & Experts & Neutral & 0.32 & 0.44 & 0.41 & 0.27 \\ 
  DistilBART & Experts & Right-wing & 0.37 & 0.47 & 0.26 & 0.63 \\ 
  DeBERTa & Experts & Left-wing & 0.06 & 0.49 & 0.42 & 0.03 \\ 
  DeBERTa & Experts & Neutral & 0.43 & 0.40 & 0.43 & 0.42 \\ 
  DeBERTa & Experts & Right-wing & 0.31 & 0.41 & 0.21 & 0.55 \\ 
  DistilBERT & Experts & Left-wing & 0.35 & 0.47 & 0.42 & 0.30 \\ 
  DistilBERT & Experts & Neutral & 0.51 & 0.46 & 0.41 & 0.67 \\ 
  DistilBERT & Experts & Right-wing & 0.11 & 0.46 & 0.19 & 0.08 \\ 
  RuBERT & Experts & Left-wing & 0.45 & 0.42 & 0.39 & 0.54 \\ 
  RuBERT & Experts & Neutral & 0.20 & 0.45 & 0.39 & 0.14 \\ 
  RuBERT & Experts & Right-wing & 0.28 & 0.42 & 0.23 & 0.34 \\ 
  RoBERTa & Experts & Left-wing & 0.54 & 0.50 & 0.37 & 0.99 \\ 
  RoBERTa & Experts & Neutral & 0.00 & 0.50 & 0.33 & 0.00 \\ 
  RoBERTa & Experts & Right-wing & 0.02 & 0.49 & 0.25 & 0.01 \\ 
  DEBATE & Experts & Left-wing & 0.00 & 0.50 & 0.80 & 0.00 \\ 
  DEBATE & Experts & Neutral & 0.57 & 0.51 & 0.40 & 0.99 \\ 
  DEBATE & Experts & Right-wing & 0.12 & 0.52 & 0.67 & 0.07 \\ 
   \hline
\end{tabular}
\endgroup
\end{table}

\begin{figure}[ht!]
	\centering
	\caption{Aggregate manifesto-level correlation coefficients of zero-shot models - crowd coding.}
	\label{2fig:zeroshot_crowd_corr}
	\includegraphics[width=0.8\linewidth]{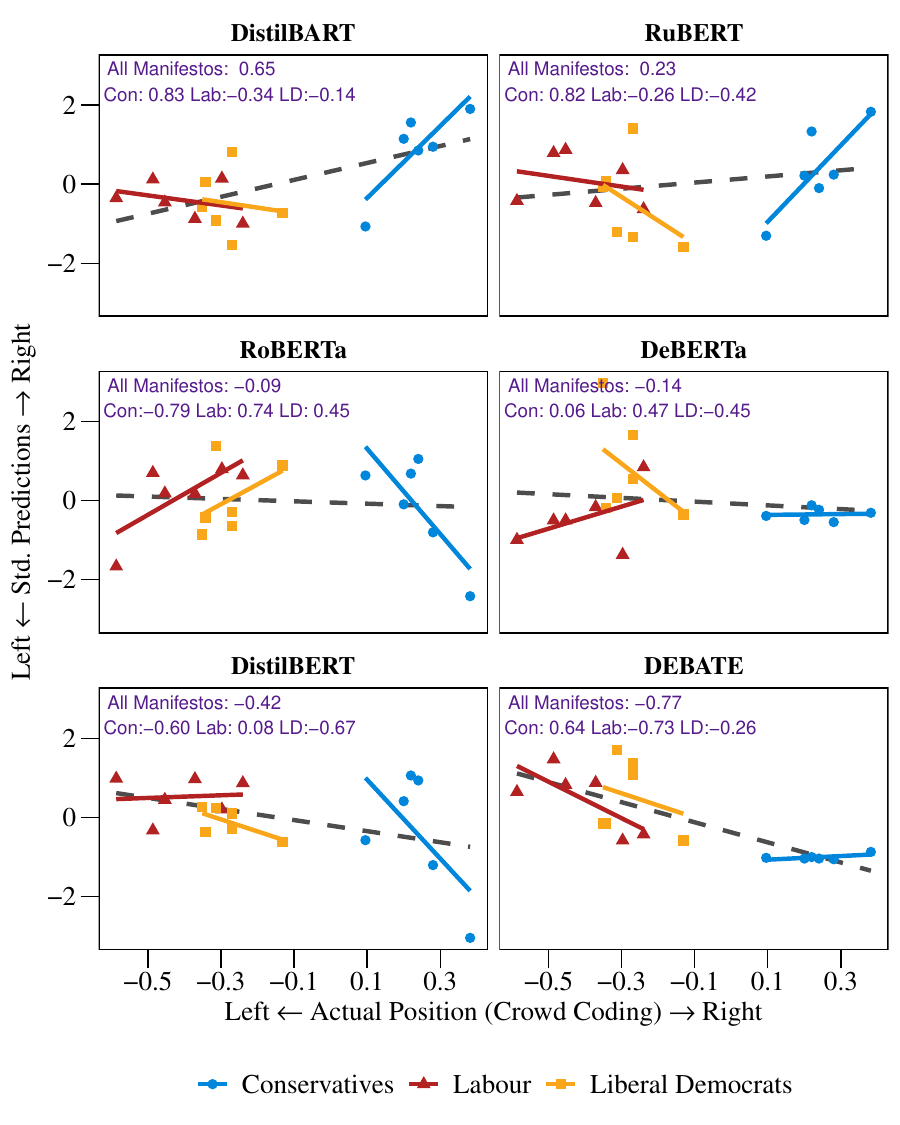}
\end{figure}
% latex table generated in R 4.3.1 by xtable 1.8-4 package
% Wed Nov  6 14:43:26 2024
\begin{table}[ht!]
\centering
\caption{Manifesto-level predicted economic ideology scores by zero-shot models.} 
\label{tab2:manif_crowd}
\begingroup\footnotesize
\resizebox{\textwidth}{!}{
\begin{tabular}{lrrrrrrrr}
  \hline
Manifesto & Crowd Pos. & Experts Pos. & DistilBART & RuBERT & DeBERTa & RoBERTa & DistilBERT & DEBATE \\ 
  \hline
Con 1987 & 0.20 & 0.39 & 1.38 & 0.94 & 1.87 & 0.31 & 0.82 & -5.34 \\ 
  Con 1992 & 0.22 & 0.36 & 1.42 & 0.98 & 2.01 & 0.39 & 0.87 & -5.19 \\ 
  Con 1997 & 0.24 & 0.37 & 1.35 & 0.93 & 1.97 & 0.43 & 0.86 & -5.34 \\ 
  Con 2001 & 0.28 & 0.54 & 1.36 & 0.94 & 1.85 & 0.24 & 0.70 & -5.41 \\ 
  Con 2005 & 0.38 & 0.39 & 1.46 & 1.00 & 1.94 & 0.08 & 0.56 & -4.67 \\ 
  Con 2010 & 0.10 & 0.23 & 1.15 & 0.88 & 1.91 & 0.39 & 0.74 & -5.27 \\ 
  Lab 1987 & -0.59 & -0.86 & 1.22 & 0.91 & 1.69 & 0.15 & 0.86 & 1.34 \\ 
  Lab 1992 & -0.49 & -0.56 & 1.27 & 0.96 & 1.87 & 0.40 & 0.76 & 4.62 \\ 
  Lab 1997 & -0.30 & -0.27 & 1.27 & 0.94 & 1.55 & 0.41 & 0.80 & -3.52 \\ 
  Lab 2001 & -0.45 & -0.43 & 1.21 & 0.96 & 1.87 & 0.34 & 0.82 & 2.04 \\ 
  Lab 2005 & -0.37 & -0.44 & 1.17 & 0.91 & 2.00 & 0.34 & 0.86 & 2.26 \\ 
  Lab 2010 & -0.24 & -0.33 & 1.15 & 0.91 & 2.37 & 0.39 & 0.85 & -2.91 \\ 
  LD 1987 & -0.31 & -0.39 & 1.16 & 0.89 & 2.08 & 0.47 & 0.81 & 5.55 \\ 
  LD 1992 & -0.35 & -0.38 & 1.20 & 0.93 & 3.16 & 0.24 & 0.81 & -1.81 \\ 
  LD 1997 & -0.27 & -0.43 & 1.34 & 0.98 & 2.26 & 0.26 & 0.79 & 3.09 \\ 
  LD 2001 & -0.34 & -0.42 & 1.26 & 0.93 & 1.99 & 0.28 & 0.76 & -1.81 \\ 
  LD 2005 & -0.13 & -0.22 & 1.18 & 0.87 & 1.93 & 0.42 & 0.74 & -3.52 \\ 
  LD 2010 & -0.27 & -0.32 & 1.10 & 0.88 & 2.67 & 0.29 & 0.76 & 4.20 \\ 
   \hline
\end{tabular}
}
\endgroup
\end{table}

% latex table generated in R 4.3.1 by xtable 1.8-4 package
% Wed Nov  6 14:48:18 2024
\begin{table}[ht!]
\centering
\caption{Comparing the performance metrics of different prompts (DistilBART zero-shot).} 
\label{tab2:compare_prompts}
\begingroup\footnotesize
\begin{tabular}{lllrrrr}
  \hline
Prompt & Source & Class & F1 & Accuracy & Prevision & Recall \\
  \hline
Prompt 1 & Crowd & Left-wing & 0.35 & 0.55 & 0.72 & 0.23 \\ 
  Prompt 1 & Crowd & Neutral & 0.02 & 0.50 & 0.13 & 0.01 \\ 
  Prompt 1 & Crowd & Right-wing & 0.55 & 0.55 & 0.39 & 0.89 \\ 
  Prompt 2 & Crowd & Left-wing & 0.34 & 0.52 & 0.68 & 0.23 \\ 
  Prompt 2 & Crowd & Neutral & 0.16 & 0.44 & 0.11 & 0.27 \\ 
  Prompt 2 & Crowd & Right-wing & 0.49 & 0.47 & 0.41 & 0.63 \\ 
  Prompt 3 & Crowd & Left-wing & 0.41 & 0.47 & 0.67 & 0.30 \\ 
  Prompt 3 & Crowd & Neutral & 0.21 & 0.52 & 0.12 & 0.78 \\ 
  Prompt 3 & Crowd & Right-wing & 0.22 & 0.47 & 0.47 & 0.14 \\ 
  Prompt 4 & Crowd & Left-wing & 0.37 & 0.45 & 0.68 & 0.26 \\ 
  Prompt 4 & Crowd & Neutral & 0.21 & 0.53 & 0.12 & 0.86 \\ 
  Prompt 4 & Crowd & Right-wing & 0.12 & 0.48 & 0.48 & 0.07 \\ 
  Prompt 1 & Experts & Left-wing & 0.33 & 0.48 & 0.52 & 0.24 \\ 
  Prompt 1 & Experts & Neutral & 0.02 & 0.50 & 0.39 & 0.01 \\ 
  Prompt 1 & Experts & Right-wing & 0.39 & 0.51 & 0.25 & 0.88 \\ 
  Prompt 2 & Experts & Left-wing & 0.31 & 0.48 & 0.47 & 0.23 \\ 
  Prompt 2 & Experts & Neutral & 0.32 & 0.44 & 0.41 & 0.27 \\ 
  Prompt 2 & Experts & Right-wing & 0.37 & 0.47 & 0.26 & 0.63 \\ 
  Prompt 3 & Experts & Left-wing & 0.40 & 0.53 & 0.52 & 0.33 \\ 
  Prompt 3 & Experts & Neutral & 0.55 & 0.52 & 0.44 & 0.73 \\ 
  Prompt 3 & Experts & Right-wing & 0.21 & 0.50 & 0.34 & 0.15 \\ 
  Prompt 4 & Experts & Left-wing & 0.38 & 0.53 & 0.53 & 0.30 \\ 
  Prompt 4 & Experts & Neutral & 0.56 & 0.52 & 0.43 & 0.81 \\ 
  Prompt 4 & Experts & Right-wing & 0.10 & 0.49 & 0.30 & 0.06 \\ 
   \hline
\end{tabular}
\endgroup
\end{table}

\begin{figure}[ht!]
	\centering
	\caption{Aggregate Manifesto-level correlation coefficients of zero-shot DistilBART using different prompt variations - crowd coding.}
	\label{2fig:prompt_crowd_corr}
	\includegraphics[width=0.8\linewidth]{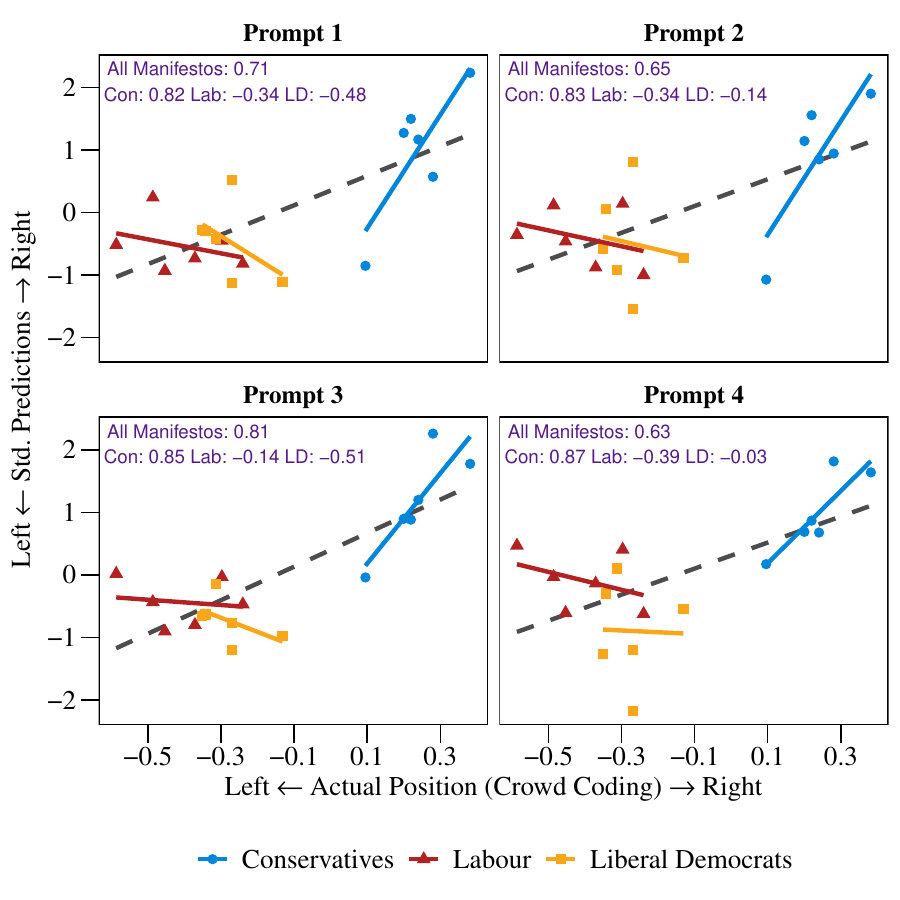}
\end{figure}
% latex table generated in R 4.3.1 by xtable 1.8-4 package
% Wed Nov  6 14:48:20 2024
\begin{table}[ht!]
\centering
\caption{Manifesto-level predicted economic ideology scores using different prompts (DistilBART model).} 
\label{tab2:prompt_crowd}
\begingroup\footnotesize
\begin{tabular}{lrrrrrr}
  \hline
Manifesto & Crowd Pos. & Experts Pos. & Prompt 1 & Prompt 2 & Prompt 3 & Prompt 4 \\ 
  \hline
Con 1987 & 0.20 & 0.39 & 1.49 & 1.38 & 0.94 & 0.80 \\ 
  Con 1992 & 0.22 & 0.36 & 1.51 & 1.42 & 0.94 & 0.83 \\ 
  Con 1997 & 0.24 & 0.37 & 1.48 & 1.35 & 0.97 & 0.80 \\ 
  Con 2001 & 0.28 & 0.54 & 1.41 & 1.36 & 1.08 & 0.96 \\ 
  Con 2005 & 0.38 & 0.39 & 1.60 & 1.46 & 1.03 & 0.94 \\ 
  Con 2010 & 0.10 & 0.23 & 1.25 & 1.15 & 0.85 & 0.72 \\ 
  Lab 1987 & -0.59 & -0.86 & 1.29 & 1.22 & 0.86 & 0.77 \\ 
  Lab 1992 & -0.49 & -0.56 & 1.37 & 1.27 & 0.81 & 0.69 \\ 
  Lab 1997 & -0.30 & -0.27 & 1.30 & 1.27 & 0.85 & 0.76 \\ 
  Lab 2001 & -0.45 & -0.43 & 1.24 & 1.21 & 0.76 & 0.61 \\ 
  Lab 2005 & -0.37 & -0.44 & 1.26 & 1.17 & 0.77 & 0.68 \\ 
  Lab 2010 & -0.24 & -0.33 & 1.26 & 1.15 & 0.81 & 0.61 \\ 
  LD 1987 & -0.31 & -0.39 & 1.30 & 1.16 & 0.84 & 0.71 \\ 
  LD 1992 & -0.35 & -0.38 & 1.32 & 1.20 & 0.79 & 0.51 \\ 
  LD 1997 & -0.27 & -0.43 & 1.40 & 1.34 & 0.78 & 0.52 \\ 
  LD 2001 & -0.34 & -0.42 & 1.31 & 1.26 & 0.79 & 0.65 \\ 
  LD 2005 & -0.13 & -0.22 & 1.22 & 1.18 & 0.76 & 0.62 \\ 
  LD 2010 & -0.27 & -0.32 & 1.22 & 1.10 & 0.73 & 0.38 \\ 
   \hline
\end{tabular}
\endgroup
\end{table}

\end{document}